\documentclass[10pt,twocolumn,letterpaper]{article}

\usepackage{iccv}              
\usepackage{comment}
\usepackage[normalem]{ulem}
\usepackage{bm}

\definecolor{iccvblue}{rgb}{0.21,0.49,0.74}
\usepackage[pagebackref,breaklinks,colorlinks,allcolors=iccvblue]{hyperref}
\usepackage{soul} 
\usepackage{xcolor} 
\usepackage{multirow}
\usepackage{pifont}
\usepackage{amssymb}
\usepackage{amsmath}
\usepackage[accsupp]{axessibility}
\definecolor{lightgreen}{rgb}{0.85, 1, 0.85}
\definecolor{lightyellow}{rgb}{1, 1, 0.8}
\definecolor{lightblue}{rgb}{0.85, 1, 1}

\newcommand{\cmark}{\ding{51}}

\newcommand{\xmark}{\ding{55}}

\def\paperID{1692} 
\def\confName{ICCV}
\def\confYear{2025}

\title{SVG-Head: Hybrid Surface-Volumetric Gaussians for High-Fidelity \\ Head Reconstruction and Real-Time Editing}

\author{
Heyi Sun$^{1}$ \quad
Cong Wang$^{1}$ \quad
Tian-Xing Xu$^{1}$ \quad \\
Jingwei Huang$^{2}$ \quad
Di Kang$^{2}$ \quad
Chunchao Guo$^{2}$ \quad
Song-Hai Zhang$^{1\dag}$\\
\\
$^{1}$ Tsinghua University\;
$^{2}$ Tencent Hunyuan
\\
\small \texttt{project page: \href{https://heyy-sun.github.io/SVG-Head/}{https://heyy-sun.github.io/SVG-Head/}}
\\
}

\definecolor{ao}{rgb}{0.0, 0.0, 1.0}
\definecolor{airforceblue}{rgb}{0.36, 0.54, 0.66}
\definecolor{ceruleanblue}{rgb}{0.16, 0.32, 0.75}
\definecolor{cerulean}{rgb}{0.0, 0.48, 0.65}
\definecolor{celestialblue}{rgb}{0.29, 0.59, 0.82}
\definecolor{azure(colorwheel)}{rgb}{0.0, 0.5, 1.0}
\definecolor{babyblue}{rgb}{0.54, 0.81, 0.94}
\definecolor{babyblueeyes}{rgb}{0.63, 0.79, 0.95}
\definecolor{ballblue}{rgb}{0.13, 0.67, 0.8}

\definecolor{asparagus}{rgb}{0.53, 0.66, 0.42}
\definecolor{ao(english)}{rgb}{0.0, 0.5, 0.0}
\definecolor{applegreen}{rgb}{0.55, 0.71, 0.0}
\definecolor{armygreen}{rgb}{0.29, 0.33, 0.13}
\definecolor{gray-asparagus}{rgb}{0.27, 0.35, 0.27}
\definecolor{green(ryb)}{rgb}{0.4, 0.69, 0.2}

\definecolor{amethyst}{rgb}{0.6, 0.4, 0.8}
\definecolor{antiquefuchsia}{rgb}{0.57, 0.36, 0.51}
\definecolor{blue-violet}{rgb}{0.54, 0.17, 0.89}
\definecolor{brightlavender}{rgb}{0.75, 0.58, 0.89}
\definecolor{brightube}{rgb}{0.82, 0.62, 0.91}
\definecolor{brilliantlavender}{rgb}{0.96, 0.73, 1.0}

\definecolor{amber}{rgb}{1.0, 0.75, 0.0}
\definecolor{amber(sae/ece)}{rgb}{1.0, 0.49, 0.0}
\definecolor{atomictangerine}{rgb}{1.0, 0.6, 0.4}
\definecolor{burntorange}{rgb}{0.8, 0.33, 0.0}
\definecolor{burntsienna}{rgb}{0.91, 0.45, 0.32}
\definecolor{cadmiumorange}{rgb}{0.93, 0.53, 0.18}
\definecolor{carrotorange}{rgb}{0.93, 0.57, 0.13}
\definecolor{chocolate(web)}{rgb}{0.82, 0.41, 0.12}
\definecolor{chromeyellow}{rgb}{1.0, 0.65, 0.0}
\definecolor{darkorange}{rgb}{1.0, 0.55, 0.0}
\definecolor{darktangerine}{rgb}{1.0, 0.66, 0.07}
\definecolor{deepcarrotorange}{rgb}{0.91, 0.41, 0.17}
\definecolor{deepsaffron}{rgb}{1.0, 0.6, 0.2}
\definecolor{fulvous}{rgb}{0.86, 0.52, 0.0}

\newcommand{\name}{{SVG-Head}}
\newcommand{\fullname}{{Surface-Volumetric Gaussian Head Avatar}}

\newcommand{\supp}{\textbf{\textcolor{olive}{Supp. Mat.}}}

\newcommand{\bfskip}{1.2mm}

\begin{document}
\maketitle

\renewcommand{\thefootnote}{}
\footnotetext[0]{ $^\dag$Corresponding authors.}

\begin{abstract}
Creating high-fidelity and editable head avatars is a pivotal challenge in computer vision and graphics, boosting many AR/VR applications. 
While recent advancements have achieved photorealistic renderings and plausible animation, head editing, especially real-time appearance editing, remains challenging due to the implicit representation and entangled modeling of the geometry and global appearance.
To address this, we propose \uline{S}urface-\uline{V}olumetric \uline{G}aussian \uline{Head} Avatar (\name), a novel hybrid representation that explicitly models the geometry with 3D Gaussians bound on a FLAME mesh and leverages disentangled texture images to capture the global appearance.
Technically, it contains two types of Gaussians, in which surface Gaussians explicitly model the appearance of head avatars using learnable texture images, facilitating real-time texture editing, while volumetric Gaussians enhance the reconstruction quality of non-Lambertian regions (\eg, lips and hair).
To model the correspondence between 3D world and texture space, we provide a mesh-aware Gaussian UV mapping method, which leverages UV coordinates given by the FLAME mesh to obtain sharp texture images and real-time rendering speed.
A hierarchical optimization strategy is further designed to pursue the optimal performance in both reconstruction quality and editing flexibility.
Experiments on the NeRSemble dataset show that \name\ not only generates high-fidelity rendering results, but also is the first method to obtain explicit texture images for Gaussian head avatars and support real-time appearance editing.
\end{abstract}
\section{Introduction}
\label{sec:intro}

Reconstructing head avatars for high-fidelity rendering, precise animation, and real-time editing remains a core challenge in computer vision and graphics, with applications in gaming~\cite{waggoner2009my}, virtual reality~\cite{DBLP:conf/ismar/HeDP20, DBLP:conf/uist/Orts-EscolanoRF16, DBLP:journals/tog/LombardiSSS18}, and film production~\cite{cogley2024digital, DBLP:journals/scpe/ZhangP24}.

Since mesh-based methods~\cite{DBLP:journals/tog/IchimBP15, DBLP:journals/tog/HuSWNSFSSCL17, DBLP:journals/tog/BaoLCZWZKHJWYZ22} rely heavily on high-quality geometry and meticulous UV coordinates for head avatar modeling, recent advances employ NeRF~\cite{nerf-DBLP:conf/eccv/MildenhallSTBRN20} or 3D Gaussian Splatting (3DGS)~\cite{3dgs-DBLP:journals/tog/KerblKLD23} to capture high-fidelity appearance from multi-view videos, with deformation fields~\cite{nerf-bs-DBLP:journals/tog/GaoZXHGZ22, pointavatar-DBLP:conf/cvpr/ZhengYWBH23, gha-DBLP:conf/cvpr/XuCL00ZL24, imavatar-DBLP:conf/cvpr/ZhengABCBH22} or rigging techniques~\cite{gaussianavatars-DBLP:conf/cvpr/QianKS0GN24} for animation. However, the implicit representation of NeRF and the entanglement of geometry and appearance for 3DGS pose challenges for editing. Despite the potential for fast, high-quality synthesis and editing~\cite{DBLP:journals/cvm/WuYZYCYG24} of 3DGS, it lacks disentangled appearance modeling, limiting avatar editability. To address this, GaussianAvatar-Editor~\cite{gaussianavatar-editor-DBLP:journals/corr/abs-2501-09978} aligns head avatars with edited images from text-to-image models~\cite{sd-DBLP:conf/cvpr/RombachBLEO22, pix2pix-DBLP:conf/cvpr/BrooksHE23}, thereby struggling to support fine-grained edits and 3D consistency.
MeGA~\cite{mega-DBLP:journals/corr/abs-2404-19026} adopts neural textures~\cite{neural-tex-DBLP:journals/tog/ThiesZN19} for high-fidelity renderings, taking minutes even hours to perform optimization-based editing. 
Due to the lack of explicit global appearance modeling, \emph{the editability of reconstructed head avatars, especially real-time texture editing, remains a critical yet long-standing problem.}

To achieve real-time texture-space editing in 3DGS while preserving photorealistic quality, we propose \fullname\ (\name), a dual-representation framework combining surface and volumetric Gaussians. Surface Gaussians (surf-GS) support texture editing, while volumetric Gaussians (vol-GS) enhance reconstruction quality.
Both types of Gaussians are bound to a FLAME mesh~\cite{flame-DBLP:journals/tog/LiBBL017} and driven by FLAME parameters for head animation.
To enable real-time editing, we first propose surf-GS that disentangle global appearance from head geometry and support sampling Gaussians' colors from two learnable texture images with our proposed mesh-aware Gaussian UV mapping, including a diffuse texture image and an expression-dependent dynamic texture image to capture dynamic facial details (\eg, wrinkles).
To avoid texture blurring caused by inconsistent UV coordinates,
surf-GS align with mesh normals and stay restricted to the mesh surface.
For further enhancing reconstruction quality of non-Lambertian regions (\eg, lips and hair), vol-GS complement residual regions where surf-GS exhibit underfitting.
Vol-GS are allowed to move around the mesh and store colors per-point to maximize their expressiveness. 
In order to effectively deliver these two types of disentangled Gaussians,
we also propose a hierarchical optimization strategy.
Specifically, surf-GS are first individually optimized to obtain coarse results and sharp texture images, followed by joint optimization with vol-GS.
A regularization loss ensures robust disentanglement of the two types.

Our contributions are summarized as follows: 

\begin{itemize}
    \item We propose a hybrid surface-volumetric Gaussian representation for animatable head creation, which utilizes surf-GS and vol-GS for high-fidelity rendering and appearance editing.
    \item We design a mesh-aware Gaussian UV mapping to model the correspondence between 3D world and texture space with low computational cost. 
    \item Experimental results on the NeRSemble dataset demonstrate that our approach achieves high-quality renderings, and is the first method to obtain explicit texture images for real-time Gaussian head avatar editing.
\end{itemize}
\section{Related Work}
\label{sec:relatedwork}

\subsection{Animatable Head Avatars}

Early mesh-based methods require a topological consistent, morphable model for reconstruction, thereby struggling to capture fine-grained geometric details (\eg hair)~\cite{DBLP:conf/icmcs/LeungTSHC00, DBLP:journals/tcsv/FuLHD08, DBLP:journals/tog/IchimBP15, DBLP:journals/tog/LombardiSSS18, DBLP:conf/fplay/HogueGJ07}.
To improve expressiveness, the Sphere Face Model~\cite{DBLP:journals/cvm/JiangJZZZTT23} introduces a hypersphere manifold latent space and combines 2D and 3D supervision for animatable head modeling. However, these models lack an explicit global appearance representation, limiting their suitability for texture editing or photorealistic rendering under challenging conditions.
Several recent methods introduce NeRF~\cite{nerf-DBLP:conf/eccv/MildenhallSTBRN20} or 3D Gaussian Splatting~\cite{3dgs-DBLP:journals/tog/KerblKLD23} to head avatars, achieving high-fidelity rendering. Due to the implicit representation of NeRF, NerFace~\cite{nerface-DBLP:conf/cvpr/GafniTZN21} uses deformation fields for animating head avatars, but suffers from noticeable floaters and struggles with precise target expressions.
NHA~\cite{nha-DBLP:conf/cvpr/GrassalPLRNT22} extends mesh-based methods and employs neural networks to improve appearance modeling.
PointAvatar~\cite{pointavatar-DBLP:conf/cvpr/ZhengYWBH23} uses colored point clouds for animatable head creation, often requiring more memory for satisfactory results.
Wang et al.~\cite{npva-DBLP:conf/siggrapha/WangKCBSZ23} propose a hybrid NPVA representation with guided surfaces for accurate expression modeling and neural volume rendering for high-quality renderings.
GaussianAvatars~\cite{gaussianavatars-DBLP:conf/cvpr/QianKS0GN24} first adapts 3DGS representations to support head avatar creation by rigging Gaussians to tracked FLAME meshes, achieving high-quality renderings and animation.
MeGA~\cite{mega-DBLP:journals/corr/abs-2404-19026} further propose a hybrid representation of mesh and Gaussians to model different head parts using more appropriate representations, obtaining higher-quality renderings and providing a way to reach head editing through optimizing for several minutes.
HERA~\cite{DBLP:conf/cvpr/CaiXWLGFGZ25} further integrates textured meshes with 3DGS in a unified rendering pipeline, employing a stable depth sorting strategy to improve visual fidelity and temporal stability, but it does not address the problem of real-time editable texture manipulation.
However, due to the lack of disentangled and explicit appearance modeling, none of above methods support real-time appearance editing.
In contrast, by representing the global appearance of head avatars as texture images, our approach supports real-time appearance editing.

\subsection{Editable 3D Representations}

Editable 3D representations serve as fundamental tools for diverse graphic applications.
Due to the inherent limitation of mesh-based methods~\cite{vhap-qian2024versatile, flame-DBLP:journals/tog/LiBBL017, DBLP:journals/tog/BaoLCZWZKHJWYZ22, DBLP:journals/tog/HuSWNSFSSCL17, DBLP:journals/tog/IchimBP15} for capturing fine-grained geometry, recent works focus on editing NeRF~\cite{nerf-DBLP:conf/eccv/MildenhallSTBRN20} or 3DGS~\cite{3dgs-DBLP:journals/tog/KerblKLD23}, which draw increasing attention for its capabilities of high-fidelity reconstruction.
However, due to their implicit and entangled modeling of global appearance, both of them struggle to perform interactive editing.
Bunch of methods~\cite{texturegs-DBLP:conf/eccv/XuHLSZ24, proteus-nerf-DBLP:journals/pacmcgit/WangDM24, ngf-DBLP:conf/iclr/ZhanLKT23, neutex-DBLP:conf/cvpr/XiangXHHS021} have been designed to enhance their ability in editing.
However, these methods focus only on the performance in novel-view synthesis and editing static scenes, which cannot be used for animatable head avatar creation.
To address this, our method binds 3D Gaussian within the local frame of triangles in a FLAME mesh and employs a mesh-aware Gaussian UV mapping, thereby achieving precise animating and editable head avatar reconstruction.

\vspace{-1mm}
\section{Method}
\label{sec:method}

\begin{figure*}
    \centering
    \includegraphics[width=\linewidth]{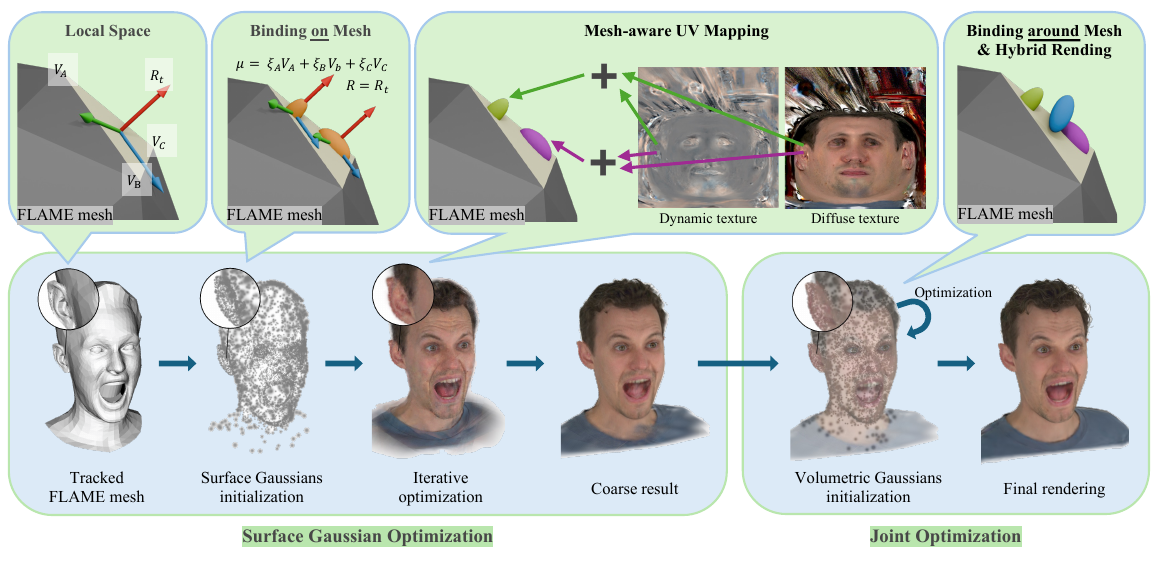}
    \vspace{-8mm}
    \caption{\textbf{Hybrid Surface-Volumetric Gaussian Head Avatar.} Our proposed {\name} consists of two types of 3D Gaussians: surface Gaussians (surf-GS) for modeling global appearance as a texture image, and volumetric Gaussians (vol-GS) for high-fidelity rendering of non-Lambertian regions. Surf-GS are placed on the mesh and use a novel mesh-aware UV mapping to fetch colors from two learnable texture images. After optimizing surf-GS, vol-GS are added around the mesh to capture geometric details for complementary modeling. This hierarchical optimization and differential hybrid rendering ensure optimal texture editing and rendering quality.
    }
    \vspace{-5mm}
    \label{fig:pipeline}
\end{figure*}



%



Given multi-view videos as input, our goal is to reconstruct a 3D head avatar supporting photorealistic rendering, precise animation and real-time appearance editing with surf-GS and vol-GS.
As shown in~\cref{fig:pipeline}, we initialize surf-GS centers (\cref{subsubsec:surfacegaussians}) on the mesh surface using random barycentric coordinates. During optimization, surf-GS can move within their parent faces without offsetting along mesh normals.
To support real-time editing, we assign two learnable texture images for appearance modeling, and a novel mesh-aware Gaussian UV mapping method to allow surf-GS to sample colors from the learned texture images.
Based on the fine-tuned result of surf-GS, we introduce vol-GS around the mesh for complementarily modeling residual regions where surf-GS exhibit underfitting (\cref{subsubsec:volumetricgaussians}).
Vol-GS store colors at each point.
To optimize this 3D representation, we employ a differentiable hybrid rendering (\cref{subsubsec:hybridrendering}) and a hierarchical optimization strategy (\cref{subsec:strategy}) for optimal editing and rendering quality.

\vspace{-1mm}
\subsection{Preliminary}

3D Gaussian Splatting (3DGS)~\cite{3dgs-DBLP:journals/tog/KerblKLD23} represents the reconstructed scene as a collection of 3D Gaussian $\{\mathcal{G}_i(x)\}_{i=1}^N$ with center position $\bm{\mu}_i\in\mathbb{R}^3$ and covariance matrix $\bm{\Sigma_i}\in \mathbb{R}^{3\times 3}$, given by 
\begin{equation}
\textstyle{
\mathcal{G}_i(\bm x) = \exp(-\frac{1}{2}(\bm x-\bm \mu_i)^T\bm \Sigma_i^{-1}(\bm x-\bm \mu_i))
}
\end{equation}

In practice, the covariance matrix $\bm \Sigma_i$ is decomposed into a rotation matrix $\bm R_i\in \mathbb R^{3\times 3}$ and a anisotropic scale factor $\bm{s}_i\in \mathbb R^3$.
To capture scene appearance, each Gaussian is assigned an opacity $o_i \in \mathbb{R}$ to describe the influence weight and spherical harmonics (SH) coefficients $\text{SH}_i \in \mathbb{R}^{16 \times 3}$ for view-dependent colors. 
The final pixel color $p$ is obtained by aggregating the color values of depth-ordered Gaussians with alpha-blending.

Due to the binding of color attributes to individual Gaussians, 3DGS struggles to model the global appearance of 3D scenes, hindering the flexibility of appearance editing. To solve this, Texture-GS~\cite{texturegs-DBLP:conf/eccv/XuHLSZ24} proposes to capture the appearance with a hybrid representation, combining per-Gaussian SH coefficients for view-dependent appearance and a global texture image with UV mapping for view-independent appearance. Unlike 3DGS, the color of each Gaussian that contributes to the alpha-blending process of pixel $p$ replies on the intersection $I(\mathcal{G}_i, \mathbf r_p)$ between the ray $\mathbf r_p$ and a 3D Gaussian $\mathcal{G}_i$, leading to a continuous 2D texture space for downstream applications. Specifically, let $\mathcal{T}\in \mathbb{R}^{H\times W\times 3}$ denote the texture image and $\phi$ denote the efficient UV mapping process in Texture-GS, the color function is given by
\begin{equation}
\textstyle{
    \mathcal{C}(\mathcal{G}_i, \mathbf r_p) = h(\phi(I(\mathcal{G}_i, \mathbf r_p)), \mathcal{T}) + \bm c^{\text{SH}_\text{res}}_i
}
\end{equation}
where $h(\cdot, \mathcal{T})$ samples value from the texture image $\mathcal{T}$ based on UV coordinates and $\bm c^{\text{SH}_\text{res}}_i \in \mathbb R^3$ denotes the residual view-dependent RGB value precomputed from SH coefficients of high degree (\textgreater 0) and camera centers.

\subsection{\fullname}

\subsubsection{Surface Gaussians}
\label{subsubsec:surfacegaussians}

To enable precise head animation, 
we follow GaussianAvatars~\cite{gaussianavatars-DBLP:conf/cvpr/QianKS0GN24} to bind 3D Gaussians to the FLAME mesh faces. Each 3D Gaussian $\mathcal{G}_i$ is defined in the local frame of its face, with parameters including the rotation matrix $\bm{R}^l_i\in\mathbb R^{3\times3}$, the center vector $\bm \mu^l_i\in\mathbb R^3$ and anisotropic scaling $\bm s^l_i\in\mathbb R^3$. The local frame of the $j$-th triangle is defined with a rotation matrix $\bm R^t_j\in \mathbb R^{3\times 3}$ to align its X and Z-axis to the first edge and the surface normal of the triangle, and a translation $\bm \mu^t_j\in\mathbb R^3$ to set its origin as the face center.
Then the global rotation $\bm R_i$, scale $s_i$ and location $\bm \mu_i$ can be recovered as follows:
\vspace{-1mm}
\begin{align}
    \bm \mu_i &= k_j\bm R^t_j\bm \mu^l_i+\bm \mu^t_j \label{eq:ga_mu} \\
    \bm R_i &= \bm R^t_j\bm R^l_i \label{eq:ga_r} \\
    \bm s_i &= k_j\bm s^l_i \label{eq:ga_s}
\end{align}
where $k_j\in \mathbb R$ denotes the scalar parameter proportionable to the $j$-th face shape. This binding preserves the relative position and rotation in the local space while allowing the Gaussians to move with the mesh during animation. Initially, a single Gaussian $\mathcal{G}_j$ is placed at the $j$-th triangle's center, which is later densified and split to multiple Gaussians ${\mathcal{G}_i}$ during optimization. 

\vspace{\bfskip}
\noindent
\textbf{Mesh-aware Gaussian UV mapping.} 
Previous 3DGS based methods~\cite{gaussianavatars-DBLP:conf/cvpr/QianKS0GN24, gha-DBLP:conf/cvpr/XuCL00ZL24, gaussianavatar-editor-DBLP:journals/corr/abs-2501-09978} capture the appearance of head avatar by assigning color attributes to each Gaussian, hindering the flexibility of appearance editing. Inspired by Texture-GS~\cite{texturegs-DBLP:conf/eccv/XuHLSZ24}, we disentangle the global appearance of head avatars from the geometry and represent it as an explicit texture image. To model the correspondence between 2D texture space and 3D world, Texture-GS employs a learnable MLP as the UV mapping function, which is tailored for the static scene and cannot be explicitly animated by the FLAME mesh.

To support editable head avatar creation, we propose a mesh-aware Gaussian UV mapping, which maps 3D positions around Gaussians to the texture image of an animatable FLAME mesh. For each ray intersection $\mathbf{p}=I(\mathbf{r}_p,\mathcal{G}_i)$, we project it onto the FLAME mesh and retrieve color via UV mapping, enabling joint optimization of Gaussians and texture images via differentiable rendering.

We compute the UV coordinate $\phi(\mathbf{p})$ for each intersection by projecting $\mathbf{p}$ to the triangle associated with $\mathcal{G}_i$ along its face normal as $\pi(\mathbf{p})$, followed by barycentric interpolation to retrieve the UV coordinates. The corresponding color is fetched from the texture image $\mathcal{T}$ for alpha-blending. This projection and interpolation process, while computationally expensive, can be condensed into a single affine transformation for efficiency:
\begin{equation}
\label{eq:comp_uv}
\textstyle{
    \phi(I(\mathbf r_p, \mathcal G_i))=\phi(\bm \mu_i)+T(\bm \mu_i)(I(\mathbf r_p, \mathcal G_i)-\bm \mu_i)
    }
\end{equation}
Here $T(\bm \mu_i) \in \mathbb R^{3\times 3}$ is a transformation matrix at the Gaussian center $\bm \mu_i$, which can be pre-computed before rendering to address this issue. 

\vspace{-2mm}
\begin{figure}[t]
    \includegraphics[width=\linewidth]{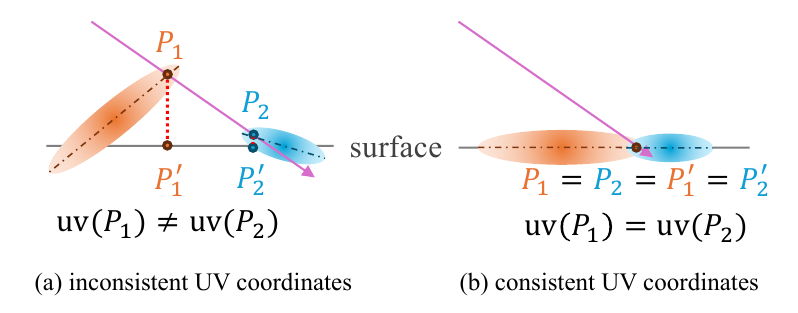}
    \vspace{-8mm}
    \caption{
    \textbf{Inconsistent UV coordinates issue.}
    (a) The intersections between a ray and multiple unconstrained Gaussians may correspond to different UV coordinates, resulting in blurred texture images.
    In contrast, (b) fixing the Gaussians' rotations and constraining them on the mesh can obtain a unique UV coordinate and sharp texture images.
    }
    \vspace{-5mm}
  \label{fig:intersection}
\end{figure}

\vspace{2.8mm}
\noindent
\textbf{Consistent UV coordinates computing.} 
\label{sec:consist_uv}

The final pixel color is the aggregate of intersection colors along ray $\mathbf{r}_p$, fetched from the texture image via mesh-aware Gaussian UV mapping. However, as shown in ~\cref{fig:intersection} (left), a ray may intersect multiple Gaussians and generate intersections corresponding to different UV coordinates, which causes blurred textures due to gradient propagation to multiple texture locations. To address this, we constrain Gaussian centers to the mesh surface and align their rotations with the surface normals.
\vspace{-1mm}
\begin{align}
    \bm \mu_i &= \xi_A \bm v_A + \xi_B \bm v_B + \xi_C \bm v_C, \\
    \bm R_i &= \bm R_j^t,
\end{align}
\vspace{-1mm}
where $\bm v_A, \bm v_B, \bm v_C\in \mathbb R^3$ are the triangle vertices and $(\xi_A, \xi_B, \xi_C) \in \mathbb R^3$ are learnable barycentric coordinates.
As shown in ~\cref{fig:intersection} (right), this binding strategy ensures intersections lie on the surface, giving each pixel a unique UV coordinate and improving texture sharpness.

\vspace{\bfskip}
\noindent
\textbf{Dynamic Texture images}
To capture dynamic facial details (\eg, wrinkles), we define the texture image $\mathcal{T}$ as the sum of two parts:
a learnable diffuse texture image $\mathcal{T}_{\text{diff}}$ for base facial colors and an expression-dependent dynamic texture $\mathcal{T}{dy}$, generated by convolutional networks conditioned on FLAME expression parameters $\psi$.

\subsubsection{Volumetric Gaussians}
\label{subsubsec:volumetricgaussians}
Since surf-GS are constrained on FLAME surface, they struggle to reconstruct complex avatars with geometry details (\eg hair). To address this, we introduce volumetric Gaussians (vol-GS) with more degrees of freedom. While vol-GS share the same geometric parameters as surf-GS (Eqs. \ref{eq:ga_mu}–\ref{eq:ga_s}), their colors are independently stored and all parameters are optimized. This gives vol-GS the flexibility of vanilla Gaussians, and they can be animated since they are bound to the FLAME mesh.

\subsubsection{Differential Hybrid Rendering}
\label{subsubsec:hybridrendering}

We propose a differential hybrid rendering module to reconcile the color formulation mechanisms of surface and volumetric Gaussians, enabling joint optimization through unified photometric supervision. To render a pixel color, a ray intersects $M$ depth-ordered Gaussians, ${\mathcal G_i}_{i=1..M}$. The color of the $i$-th Gaussian is obtained by:
\begin{equation}
\textstyle{
    \mathcal{C}(\mathcal{G}_i, \mathbf r_p)=\begin{cases}
    \bm c_i^{\text{SH}} & \text{if}\;v_i = 1 \\
    h(\phi(I(\mathbf p,\mathcal{T}_\text{dy} + \mathcal{T}_\text{diff})+\bm c_i^{\text{SH}_\text{res}}&\text{if $v_i = 0$}
    \end{cases}
    }
\end{equation}
$v_i \in \{0, 1\}$ is a binary value indicating whether $\mathcal{G}_i$ is a vol-GS ($v_i=1$) or a surf-GS ($v_i=0$).

Given the established color attributes for each intersection (\ie, $\{\mathcal{C}(\mathcal G_i, \mathbf{r}_p)\}_{i=1:M}$), the rendering incompatibility between different types of Gaussians is resolved. 
The final pixels can be directly rendered using alpha-blending. 
Furthermore, since all the aforementioned operations are differentiable, our hybrid rendering architecture allows joint optimization of both surf- and vol-GS within a unified learning framework, yielding optimal results.

\subsection{Hierarchical Optimization Strategy}
\label{subsec:strategy}

Joint optimization of both surf- and vol-GS from scratch is highly under-constrained,  where vol-GS prevents surf-GS from capturing texture details, generating suboptimal texture images and editing effects.
To address this issue, we design a hierarchical optimization strategy that divides our optimization pipeline into two stages, including surface Gaussian optimization and joint optimization.


\subsubsection{Learnable Parameters}

For convenience, we list all learnable parameters here.
The FLAME shape, expression and pose parameters are denoted as $\beta$, $\psi$, and $\varphi$, respectively.
For $i$-th surface Gaussian, $(\xi_A, \xi_B, \xi_C)$ are the barycentric coordinates used to compute its center,
$o^{(s)}_i$ and $\bm s^{(s)}$ denotes its local scaling and opacity, respectively. $\text{SH}^{(s)}_{i,\text{res}} \in \mathbb{R}^{15 \times 3}$ is the SH coefficients of high degree. $\mathcal{T}_\text{diff}$ is a learnable diffuse texture image for sampling diffuse colors, while $\mathcal{T}_\text{dy}$ is an expression-dependent dynamic texture image for modeling dynamic details.
%
For $i$-th volumetric Gaussian, $\bm \mu^{(v)}_i$, $o^{(v)}_i$, $\bm s^{(v)}_i$, $\bm R^{(v)}_i$ and $\text{SH}^{(v)}_i \in \mathbb{R}^{16 \times 3}$ denote its center, opacity value, local scaling, local rotation and SH coefficients, respectively.

\subsubsection{Surface Gaussian Optimization}

In this stage, we optimize all learnable parameters related to the FLAME mesh (\ie, $\beta$, $\psi$, and $\varphi$) and surf-GS (\ie, $\xi_A, \xi_B, \xi_C, o^{(s)}, \bm s^{(s)}, \text{SH}^{(s)}_\text{res}, \mathcal{T}_\text{diff}$ and $\mathcal{T}_\text{dy}$) with two photometric losses $\mathcal{L}_{\text{rgb}}$ and $\mathcal{L}_{\text{rgb}}^\text{diff}$, and a scaling loss $\mathcal{L}_\text{scale}$.

\vspace{\bfskip}
\noindent
\textbf{Photometric loss.} 
We take a combination of L1 and D-SSIM loss terms as the photometric supervisions, as described in 3D Gaussian Splatting~\cite{3dgs-DBLP:journals/tog/KerblKLD23}:
\begin{equation}
\label{loss:rgb1}
\textstyle{
    \mathcal{L}_{\text{rgb}}=(1-\lambda)||\bm{\hat{I}}_\text{surf} - \bm{I}||+\lambda (1 - \text{SSIM}(\bm{\hat{I}}_\text{surf}, \bm{I}) ).
}
\end{equation}
$\bm{I}$ and $\bm{\hat{I}}_\text{surf}$ are the ground truth and our surf-GS's rendering, respectively.
To further generate plausible texture maps for editing, we introduce an additional photometric loss served as regularization:
\begin{equation}
\label{loss:basergb}
\textstyle{
    \mathcal{L}_{\text{rgb}}^\text{diff}=(1-\lambda)||\bm{\hat{I}}_\text{diff} - \bm{I}||+\lambda (1 - \text{SSIM}(\bm{\hat{I}}_\text{diff}, \bm{I}) ),
    }
\end{equation}
where $\bm{\hat{I}}_\text{diff}$ is rendered only from the diffuse texture map.

\vspace{\bfskip}
\noindent
\textbf{Scaling loss.} 
Similar to GaussianAvatars~\cite{gaussianavatars-DBLP:conf/cvpr/QianKS0GN24}, we incorporate a scaling loss to avoid too large Gaussians causing wrong UV computation and jittering artifacts during animation.
The scaling regularization is defined as:
\begin{equation}
\label{loss:scaling}
\textstyle{
    \mathcal{L}_\text{scale} = \lVert \max(\bm s^{(s)},\varepsilon_\text{scale}) \rVert_2.
    }
\end{equation}
Here, $\varepsilon_\text{scale}=0.6$ is a hyperparameter.

Our total loss for this stage is formulated as follows:
\begin{equation}
\begin{aligned}
\textstyle{
\mathcal{L}_{\text{surf}}=\mathcal{L}_{\text{rgb}}+\lambda_{\text{rgb}}^\text{diff}\mathcal{L}_{\text{rgb}}^\text{diff}+\lambda_\text{scale}\mathcal{L}_\text{scale}.
}
\end{aligned}
\end{equation}

\begin{figure*}
    \centering
    \includegraphics[width=0.98\linewidth]{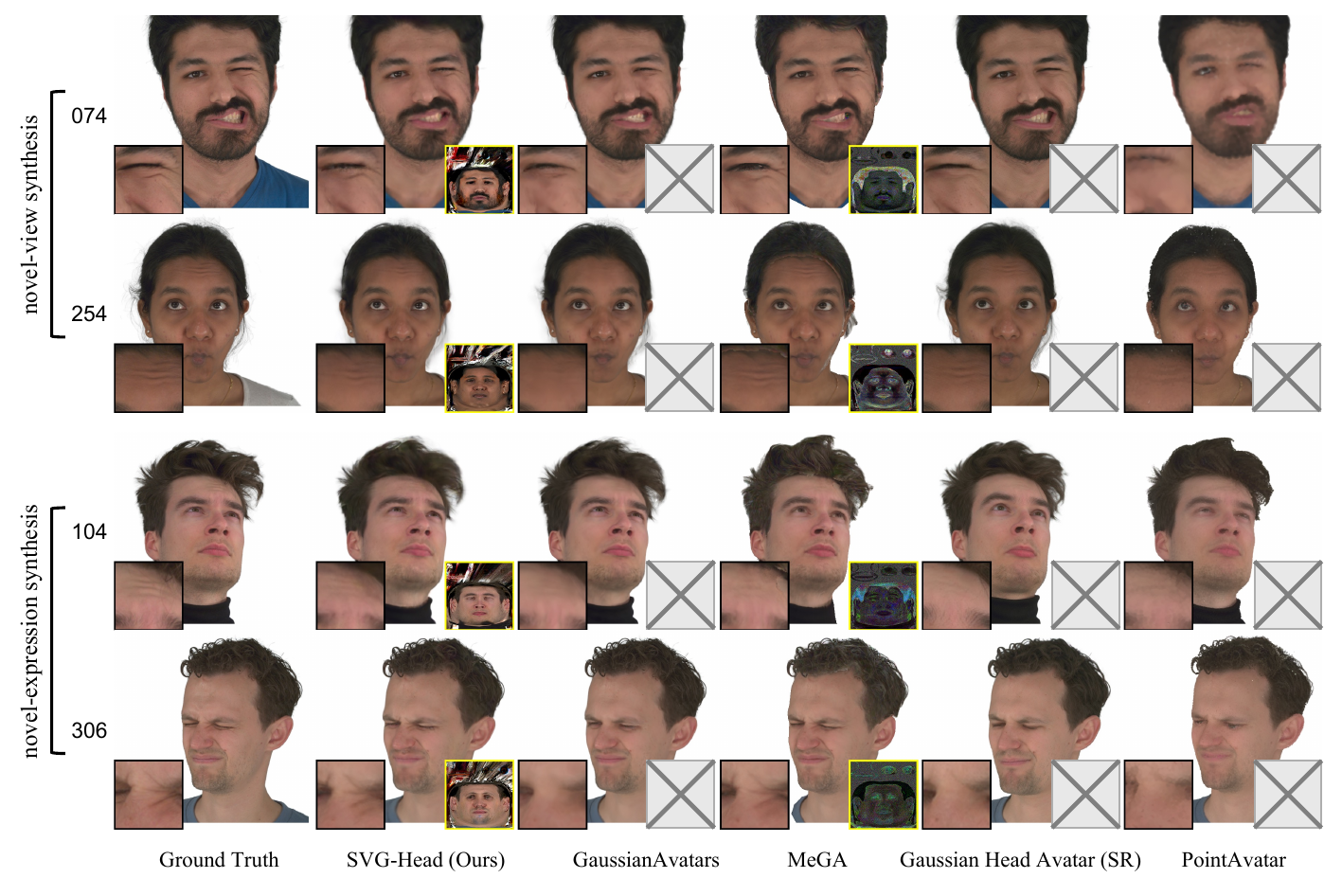}
    \vspace{-5mm}
    \caption{
    \textbf{Comparisons with State-of-the-Art Methods.}
    SVG-Head captures subtle dynamic facial details (\eg, wrinkles) and is the only method that obtains an explicit texture image for real-time editing.
    Note that Gaussian Head Avatar uses a super-resolution (SR) module.
    }
    \vspace{-5mm}
    \label{fig:results}
\end{figure*}

\subsubsection{Joint Optimization}

After the surface Gaussian optimization stage, we obtain sharp texture maps, enabling high-quality appearance editing. However, relying solely on surface Gaussians leads to suboptimal rendering quality in volumetric regions, particularly around the lips and hair. 
To address this, we introduce vol-GS and jointly optimize both types of Gaussians with a parameter scheduling scheme.
Specifically, to achieve the optimal performance in both rendering quality and editing flexibility, we selectively optimize a subset of surf-GS parameters (\ie, $o^{(s)}$ and $\mathcal{T}_\text{dy}$) and all vol-GS parameters (\ie, $\bm \mu^{(v)}$, $o^{(v)}$, $\bm s^{(v)}$, $\bm R^{(v)}$ and $\text{SH}^{(v)}$).
We also halt the densification and pruning process of the surf-GS.

In this stage, we employ a similar photometric loss $\mathcal{L}_{\text{rgb}}$ to supervise the rendering results of all Gaussians and a scaling loss $\mathcal{L}_\text{scale}$ for regularization, as mentioned before.
Two additional regularization losses for Gaussian positions $\mathcal{L}_{p}$ and alpha maps $\mathcal{L}_{a}$ are introduced.


\vspace{\bfskip}
\noindent
\textbf{Position loss.} To prevent vol-GS from drifting too far from their parent faces and impairing animation, we introduce a position regularization loss as:
\begin{equation}
\label{loss:position}
\textstyle{
    \mathcal{L}_\text{pos} = \lVert \max(\bm\mu^{(v)},\varepsilon_\text{pos}) \rVert_2,
    }
\end{equation}
where $\varepsilon_\text{pos}=1$ is a threshold used to only punish large positional errors.
%

\vspace{\bfskip}
\noindent
\textbf{Alpha loss.} To prevent the opacity of surf-GS from decreasing excessively and impairing editing effects, we introduce an alpha regularization loss applied to the renderings of surf-GS as:
%
\begin{equation}
    \textstyle{
    \mathcal{L}_{a}= ||\bm{A_s}-1||_2^2,
    }
\end{equation}
where $\bm{A_s}$ denotes the alpha map rendered using only surf-GS.

Our final loss function in this stage is:
\begin{equation}
\begin{aligned}
\textstyle{
    \mathcal{L} =\mathcal{L}_{\text{rgb}}+\lambda_a \mathcal{L}_a +\lambda_\text{scale}\mathcal{L}_\text{scale}+\lambda_\text{pos}\mathcal{L}_\text{pos},
    }
\end{aligned}
\end{equation}
\section{Experiments}
\label{sec:experiments}

We evaluate our method on the NeRSemble dataset~\cite{nersemble-DBLP:journals/tog/KirschsteinQGWN23}, which provides multi-view video recordings of multiple subjects along with calibrated camera parameters for all 16 viewpoints.
Following previous methods~\cite{gaussianavatars-DBLP:conf/cvpr/QianKS0GN24, mega-DBLP:journals/corr/abs-2404-19026}, we downsample images to a resolution of 802$\times$550, and use the train/test split as: 9 out of 10 expression sequences and 15 out of 16 available cameras for training and the remaining expression sequence and camera for evaluation.

\subsection{Head Reconstruction and Animation}

We conduct comparative experiments with PointAvatar~\cite{pointavatar-DBLP:conf/cvpr/ZhengYWBH23}, Gaussian Head Avatar with super-resolution (SR)~\cite{gha-DBLP:conf/cvpr/XuCL00ZL24}, GaussianAvatars~\cite{gaussianavatars-DBLP:conf/cvpr/QianKS0GN24}, and MeGA~\cite{mega-DBLP:journals/corr/abs-2404-19026} 
All baselines are trained from scratch using their public codes.
~\cref{tab:results} shows quantative comparisons on novel-view and novel-expression synthesis. 
Our approach achieves the best evaluation metrics among all editable reconstruction methods, while generating comparable rendering quality to those methods that do not support appearance editing and have limited applications.
Note that the reported metrics are calculated based on all image pixels and MeGA obtains rather bad metrics due to their inability to model plausible shoulder regions.
%
%
\cref{fig:results} shows qualitative results. 
Thanks to our complementary modeling and the incorporation of dynamic texture maps,
our \name\ generates high-fidelity renderings with finer-grained facial texture details (\eg, wrinkles). 
MeGA obtains plausible visual results along with a neural texture map.
However, as shown in ~\cref{fig:edit}, the neural texture map requires several minutes or even hours for large regions to achieve optimization-based editing, struggling to reach real-time appearance editing like our method.
Note that although Gaussian Head Avatar generates promising renderings for novel-view synthesis, it struggles to generate accurate target expressions due to its heavy reliance on the implicit deformation fields and super-resolution modules.
More detailed results (including videos) can be found in our \supp.

\begin{table}
  \centering
  \small
  \setlength{\tabcolsep}{3pt} 
  \caption{
  \textbf{Comparisons with State-of-the-Art Methods.}
  SVG-Head obtains the best metrics among editable reconstruction methods, as well as metrics comparable to non-editable reconstruction methods. 
  \sethlcolor{lightgreen}\hl{Green} indicates the best and \sethlcolor{lightyellow}\hl{yellow} indicates the second.
  }
  \vspace{-3mm}
  \resizebox{\linewidth}{!}{
      \begin{tabular}{@{}l|c|ccc|ccc@{}}
        \toprule
            \multirow{2}{*}{Method}& \multirow{2}{*}{Editing} & \multicolumn{3}{c|}{Novel-View Synthesis} & \multicolumn{3}{c}{Novel-Expr. Synthesis} \\
            & & PSNR$\uparrow$ & SSIM$\uparrow$ & LPIPS$\downarrow$ & PSNR$\uparrow$ & SSIM$\uparrow$ & LPIPS$\downarrow$ \\
        \midrule
        PointAvatar & \xmark & 25.8 & 0.893 & 0.097 & 23.4 & 0.884 & 0.102 \\
        Gaussian Head Avatar & \xmark & 29.5 & 0.894 & 0.084 & 22.5 & 0.853 & 0.144 \\
        GaussianAvatars & \xmark & \sethlcolor{lightgreen}\hl{31.6} & \sethlcolor{lightgreen}\hl{0.938} & \sethlcolor{lightgreen}\hl{0.065} & \sethlcolor{lightgreen}\hl{26.0} & \sethlcolor{lightgreen}\hl{0.910} & \sethlcolor{lightgreen}\hl{0.076} \\
        \midrule
        MeGA$^*$ & \cmark & 15.2 & 0.853 & 0.207 & 17.4 & 0.867 & 0.181 \\
        \name\ (Ours) & \cmark & \sethlcolor{lightyellow}\hl{30.3} & \sethlcolor{lightyellow}\hl{0.931} & \sethlcolor{lightyellow}\hl{0.078} & \sethlcolor{lightgreen}\hl{26.0} & \sethlcolor{lightgreen}\hl{0.910} & \sethlcolor{lightyellow}\hl{0.087} \\
        \bottomrule
      \end{tabular}
  }
  \vspace{-3mm}
  \label{tab:results}
\end{table}

\subsection{Real-Time Appearance Editing}

\begin{figure}
    \centering
    \includegraphics[width=\linewidth]{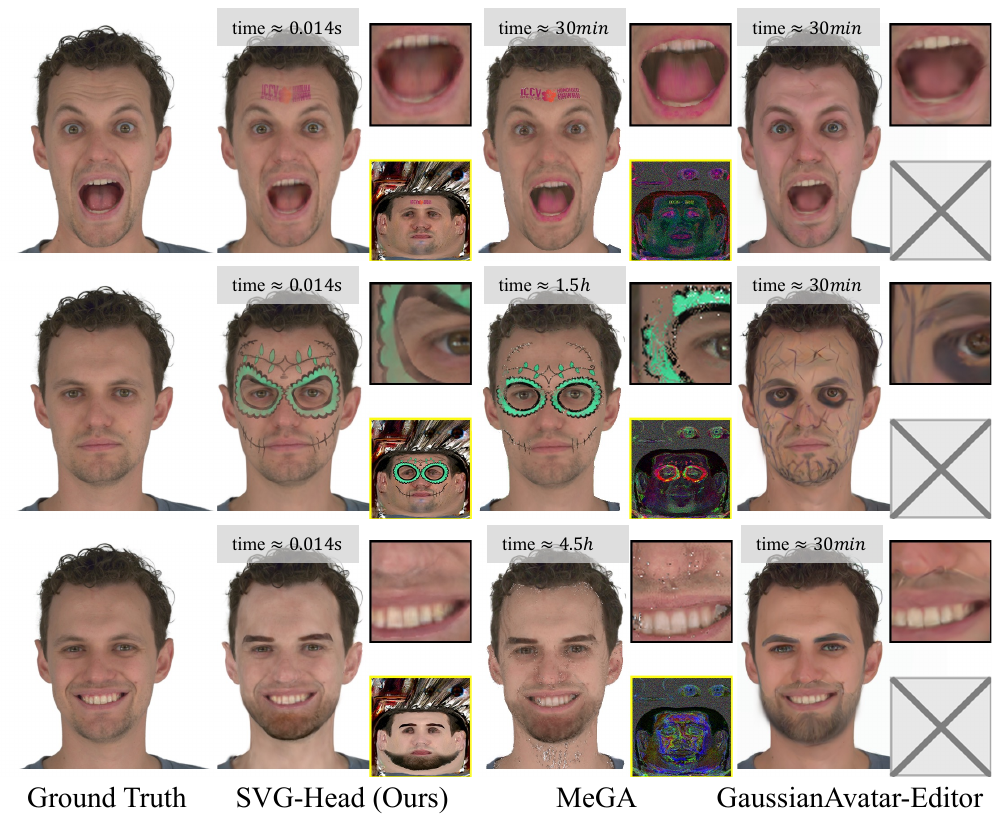}
    \vspace{-6mm}
    \caption{
    \textbf{Comparison on Editing Effects.}
    SVG-Head supports fine-grained and real-time appearance editing by simply drawing on the texture images, while MeGA and GaussianAvatar-Editor takes minutes or even hours to generate worse editing effects.
    }
    \vspace{-7mm}
    \label{fig:edit}
\end{figure}

In this section, we conduct qualitative comparisons with GaussianAvatar-Editor (GA-Editor)~\cite{gaussianavatar-editor-DBLP:journals/corr/abs-2501-09978} and MeGA~\cite{mega-DBLP:journals/corr/abs-2404-19026} to highlight our finer-grained editing capability and real-time processing speed. 

As illustrated in~\cref{fig:edit},
our approach supports finer-grained and more natural texture editing in real time by simply drawing on our disentangled and explicit diffuse texture maps.
In contrast, MeGA usually takes minutes or even hours for large regions to achieve optimization-based editing and struggles to accurately match each editing detail (\eg, missing colored painting in Row 2).
Moreover, due to the fine-tuning of the color decoder, MeGA also change colors of some regions (\eg, the mouth region in Row 1) that are not expected to be edited. 
GA-Editor also requires several minutes to achieve optimization-based editing and is extremely hard to achieve fine-grained texture editing due to taking only text prompt as editing conditions.

\subsection{Ablation Studies}

In this section, we conduct a series of ablation studies to validate the effectiveness of our designs.

\begin{table*}
  \centering
  \small
  \setlength{\tabcolsep}{3pt} 
  \caption{
  \textbf{Ablation Studies on Subject 306.}
  We demonstrate the effects of each propose component.
  (a) shows the importance of constraining surface Gaussians (surf-GS) on the mesh and fixing their rotations.
  (b) shows the positive effects of slightly fine-tuning FLAME parameters.
  (c) shows the superiority of our hierarchical optimization strategy.
  (d) shows the roles of our proposed losses.
  \sethlcolor{lightgreen}\hl{Green} indicates the best, and \sethlcolor{lightyellow}\hl{yellow} indicates the second.
  }
  \vspace{-3mm}
  \resizebox{0.85\linewidth}{!}{ 
    \begin{tabular}{l|l|c|c|c|c|ccc|ccc} 
        \toprule
         \multirow{2}{*}{Label} & \multirow{2}{*}{Name} & \multirow{2}{*}{Surf-GS}  & \multirow{2}{*}{Strategy} & \multirow{2}{*}{Losses} & \multicolumn{3}{c|}{Novel-View Synthesis} & \multicolumn{3}{c}{Novel-Expr. Synthesis} \\
         & &  & & & PSNR$\uparrow$ & SSIM$\uparrow$ & LPIPS$\downarrow$ & PSNR$\uparrow$ & SSIM$\uparrow$ & LPIPS$\downarrow$\\
        \midrule
          & \name & &  & & 33.4 & \sethlcolor{lightyellow}\hl{0.966} & \sethlcolor{lightgreen}\hl{0.037} & \sethlcolor{lightyellow}\hl{30.9} & \sethlcolor{lightgreen}\hl{0.956} & \sethlcolor{lightgreen}\hl{0.037} \\
        \midrule
        (a.1) & SVG-optR & opt. $\bm R^{(s)}$ \& on mesh & & & 32.9 & 0.964 & 0.038 & 30.0 & 0.951 & 0.040 \\
        (a.2) & SVG-optR\&P & opt. $\bm R^{(s)}$ \& around mesh &  & & \sethlcolor{lightgreen}\hl{33.8} & 0.964 & 0.040 & 29.6 & 0.946 & 0.044 \\
        \midrule
        (b.1) & SVG-freeze & & freeze all surf-GS & & 30.9 & 0.953 & 0.048 & 30.1 & 0.947 & 0.044 \\
        (b.2) & SVG-nofreeze & & opt. all surf-GS & & 30.7 & 0.948 & 0.053 & 28.4 & 0.939 & 0.051 \\
        (b.3) & SVG-onestage & & joint opt. from scratch & & 30.6 & 0.949 & 0.053 & 28.0 & 0.939 & 0.051 \\
        \midrule
        (c.1) & SVG-nodirgb & & & w/o $\mathcal{L}_{\text{rgb}}^{\text{di}}$ & 33.2 & 0.965 & 0.038 & \sethlcolor{lightgreen}\hl{31.1} & \sethlcolor{lightgreen}\hl{0.956} & 0.038 \\
        (c.2) & SVG-noalpha & & & w/o $\mathcal{L}_a$& \sethlcolor{lightyellow}\hl{33.5} & \sethlcolor{lightgreen}\hl{0.967} & \sethlcolor{lightgreen}\hl{0.037} & 30.2 & 0.955 & \sethlcolor{lightgreen}\hl{0.037} \\
        \bottomrule
    \end{tabular}
  } 
  \vspace{-5mm}
  \label{tab:ablation}
\end{table*}

\begin{figure}[t]
    \centering
    \begin{subfigure}{0.32\linewidth} 
        \includegraphics[width=\linewidth]{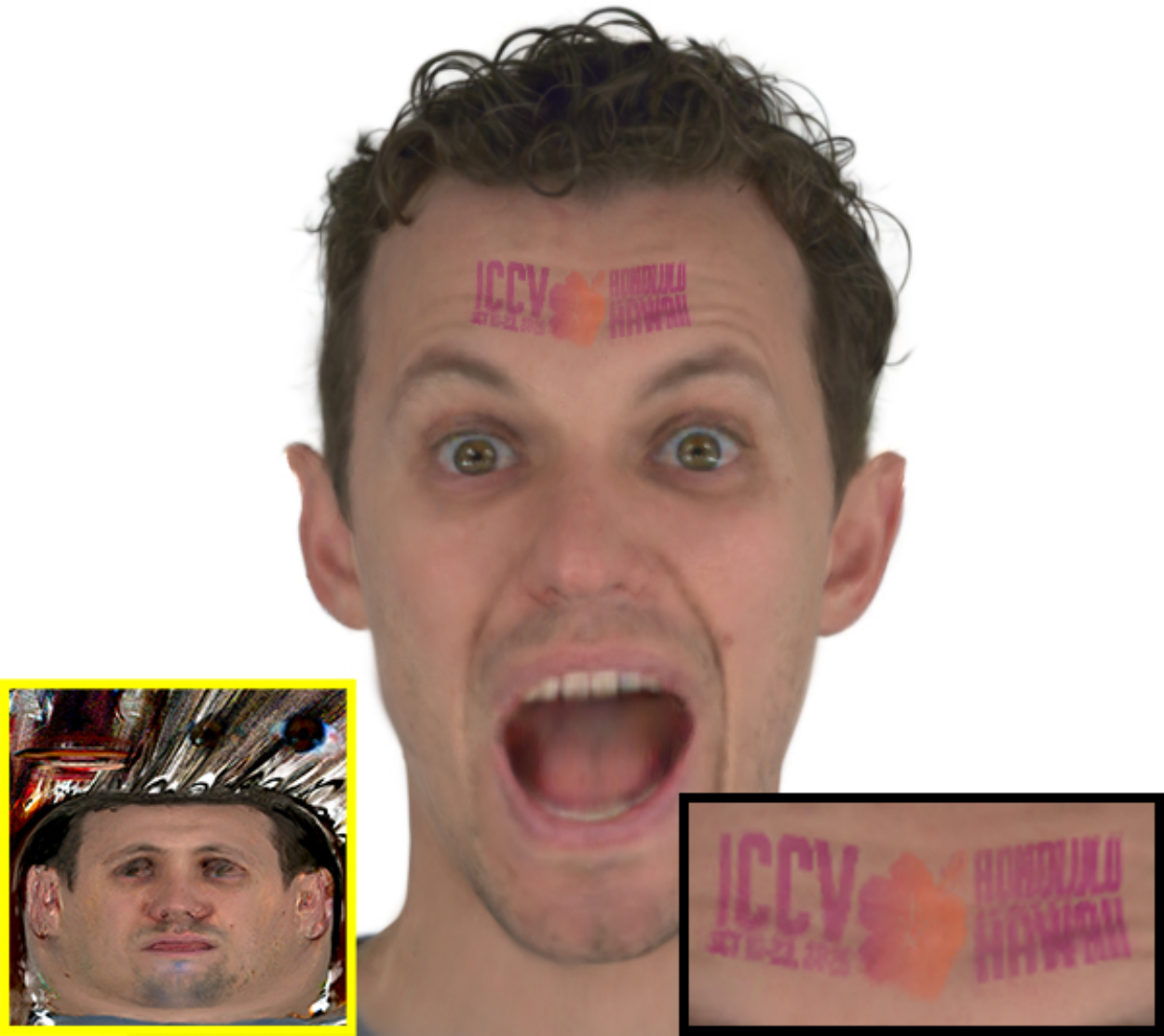}
        \caption{\name\ (Ours)}
    \end{subfigure}
    \begin{subfigure}{0.32\linewidth}
        \includegraphics[width=\linewidth]{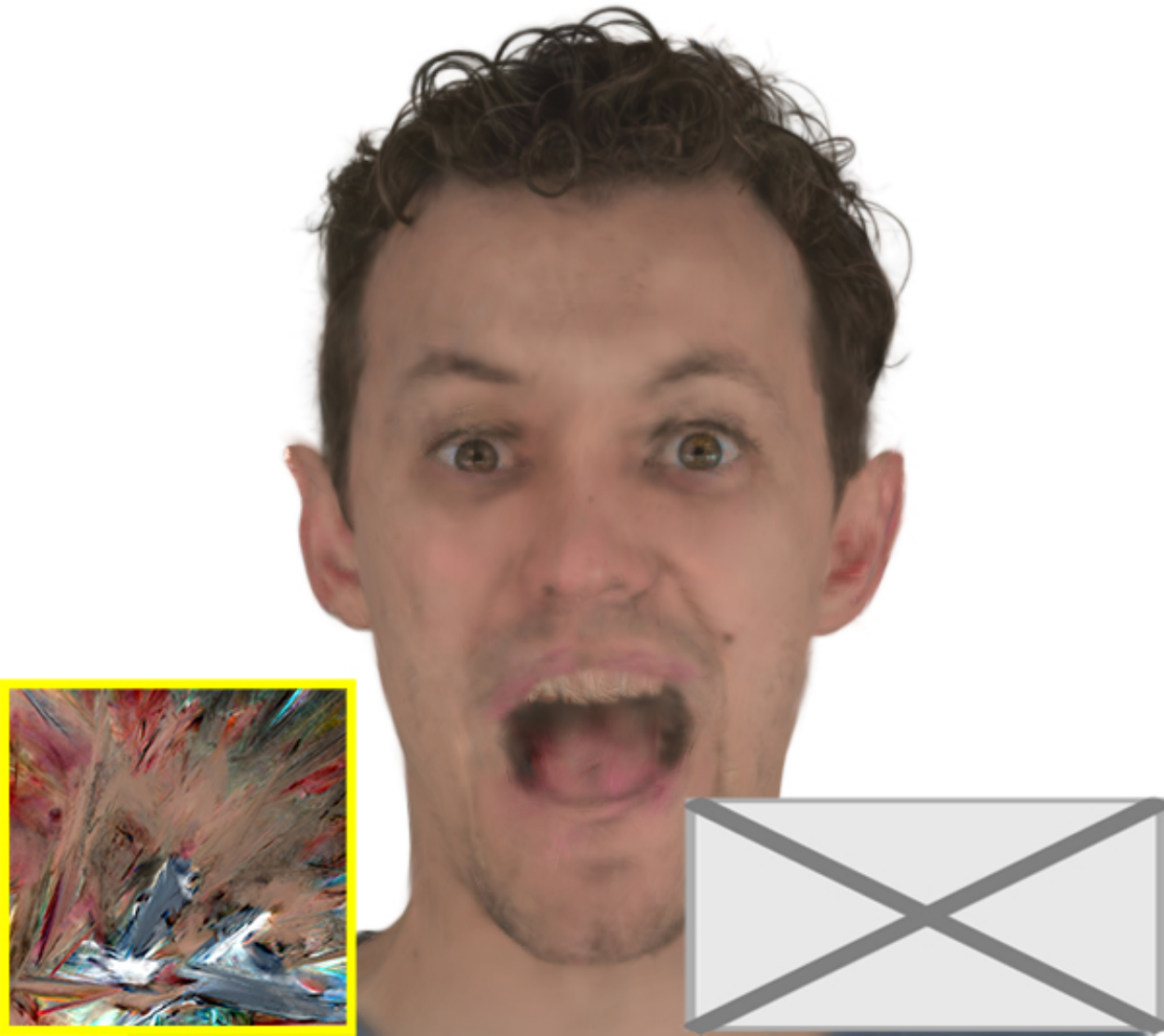}
        \caption{SVG-optR\&P}
    \end{subfigure}
    \begin{subfigure}{0.32\linewidth}
        \includegraphics[width=\linewidth]{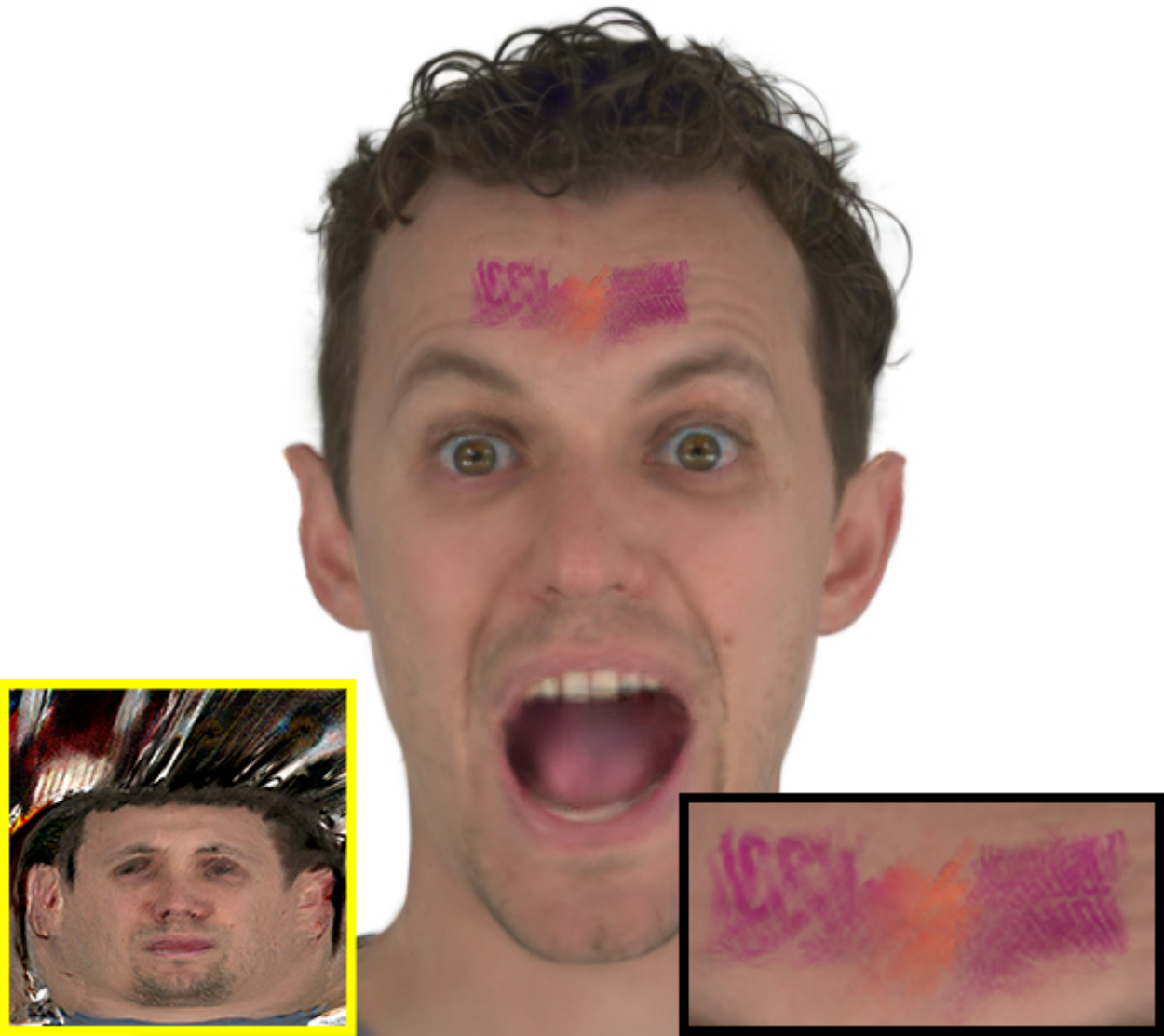}
        \caption{SVG-optR}
    \end{subfigure}
    \begin{subfigure}{0.32\linewidth}
        \includegraphics[width=\linewidth]{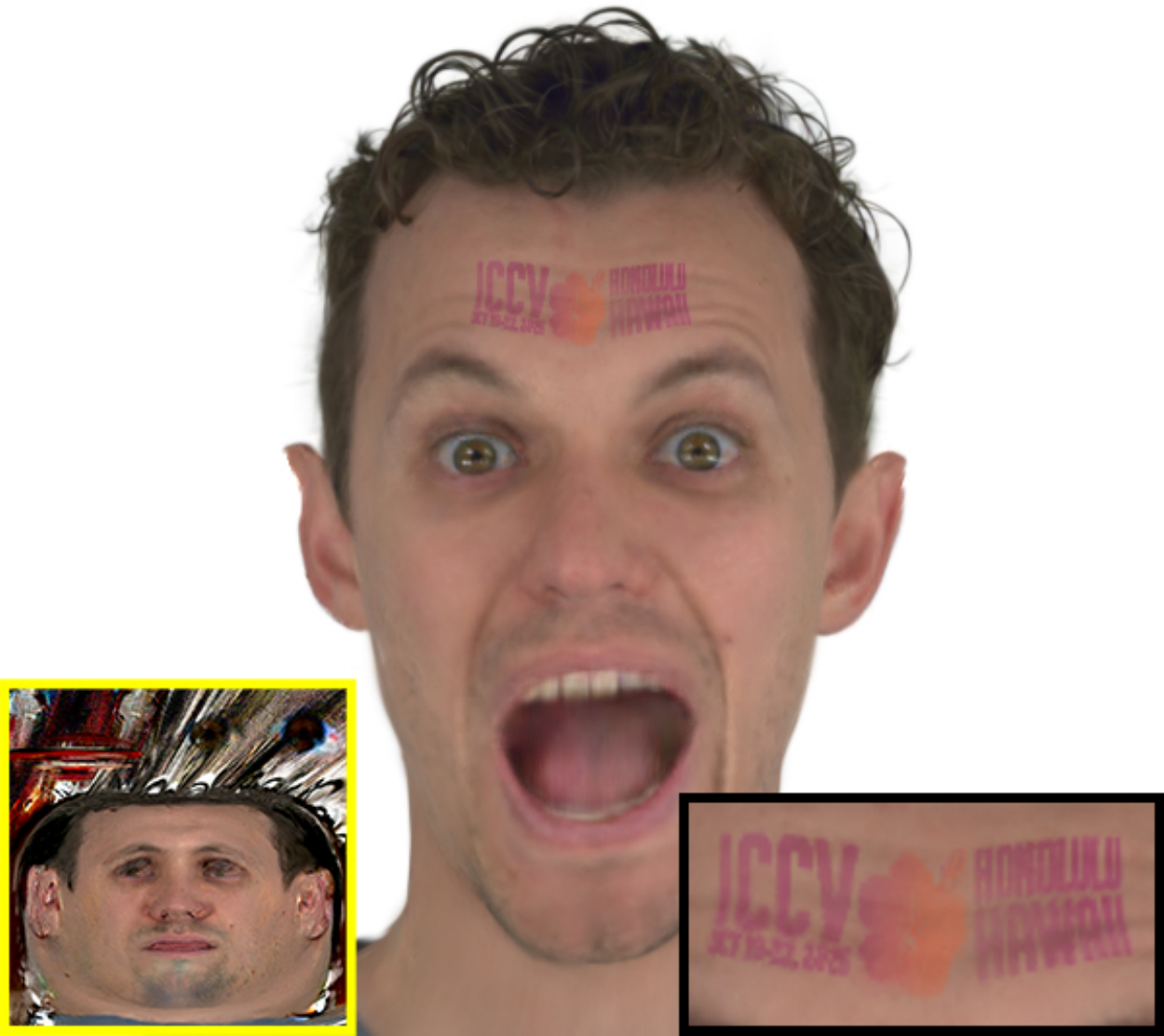}
        \caption{SVG-noalpha}
    \end{subfigure}
    \begin{subfigure}{0.32\linewidth}
        \includegraphics[width=\linewidth]{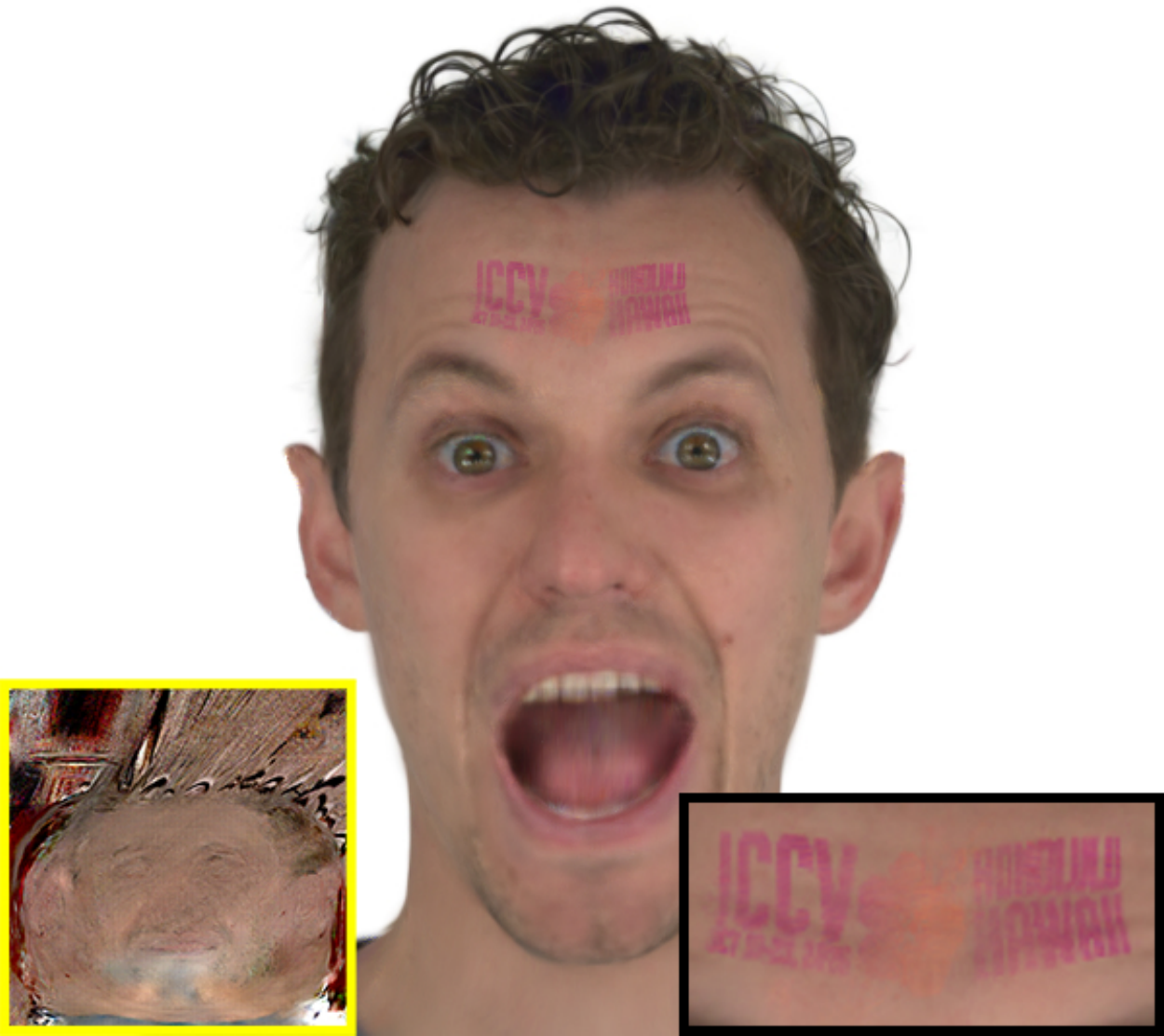}
        \caption{SVG-nodirgb}
    \end{subfigure}
    \vspace{-3mm}
    \caption{
    \textbf{Ablation Studies} on constraining surface Gaussians (surf-GS) and different losses.
    Both unconstrained surf-GS (b, c) and removing $\mathcal{L}_{\text{rgb}}^{\text{diff}}$ (e) cause a suboptimal texture image.
    Removing $\mathcal{L}_a$ (d) results in transparent editing effects.
    }
    \vspace{-5mm}
    \label{fig:ablation-edit}
\end{figure}

\vspace{\bfskip}
\noindent
\textbf{Surface Gaussians design.}
\cref{tab:ablation}(a.1)-(a.2) demonstrates the importance of constraining the surf-GS on the mesh while fixing their rotations.
``SVG-optR\&P'' allows the surf-GS to detach from the mesh and optimize the local rotations and positions like vol-GS, while ``SVG-optR'' only unfreeze the optimization of the rotations.
%
Both of them result in blurry texture images due to the inconsistent UV coordinates computing (\cref{sec:consist_uv}), impairing the rendering (30.9 vs. 30.0/29.6 PSNR) and editing quality (\cref{fig:ablation-edit}(b)-(c)).


\begin{figure}[t]
    \centering
    \begin{subfigure}{0.32\linewidth} 
        \includegraphics[width=\linewidth]{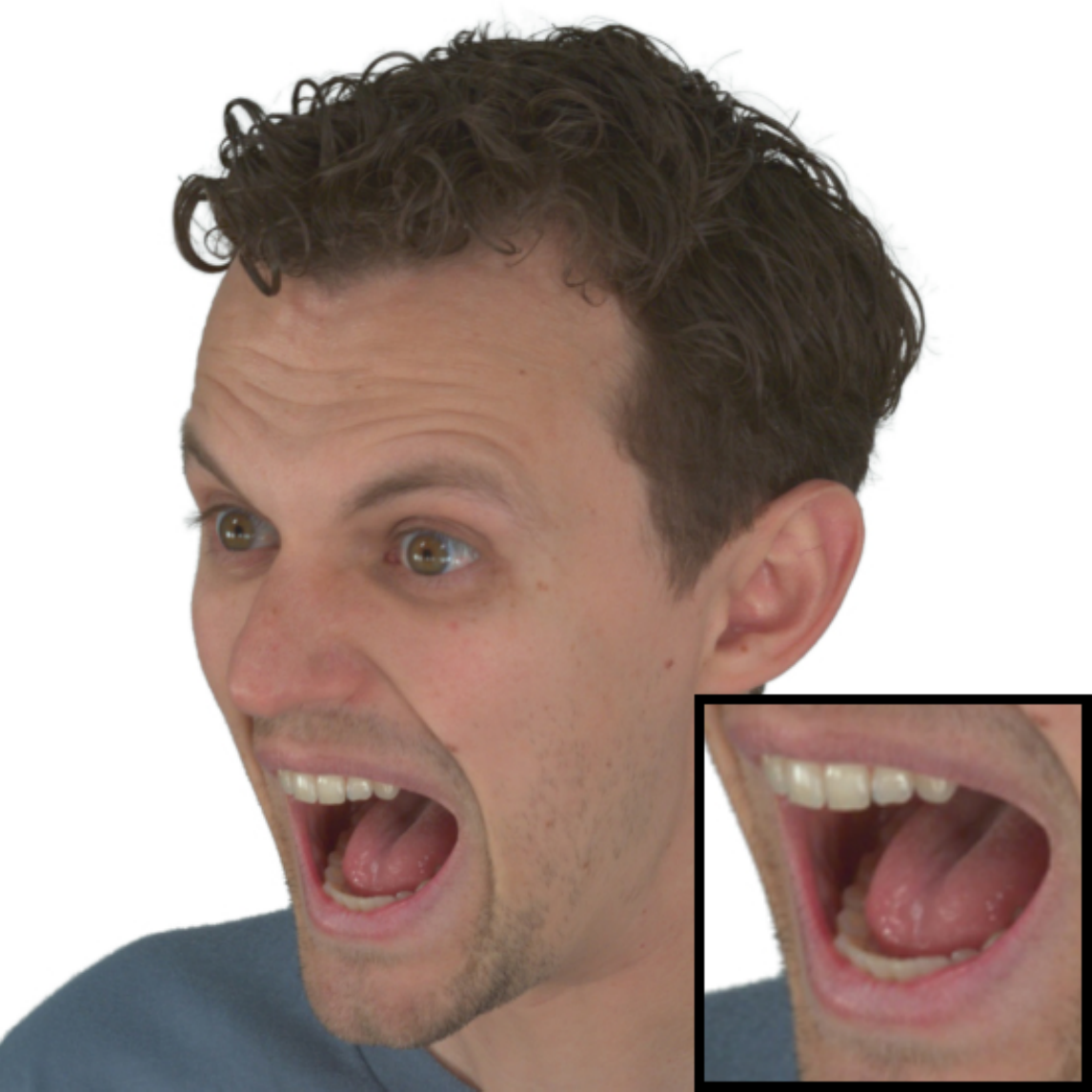}
        \caption{Ground Truth}
    \end{subfigure}
    \begin{subfigure}{0.32\linewidth} 
        \includegraphics[width=\linewidth]{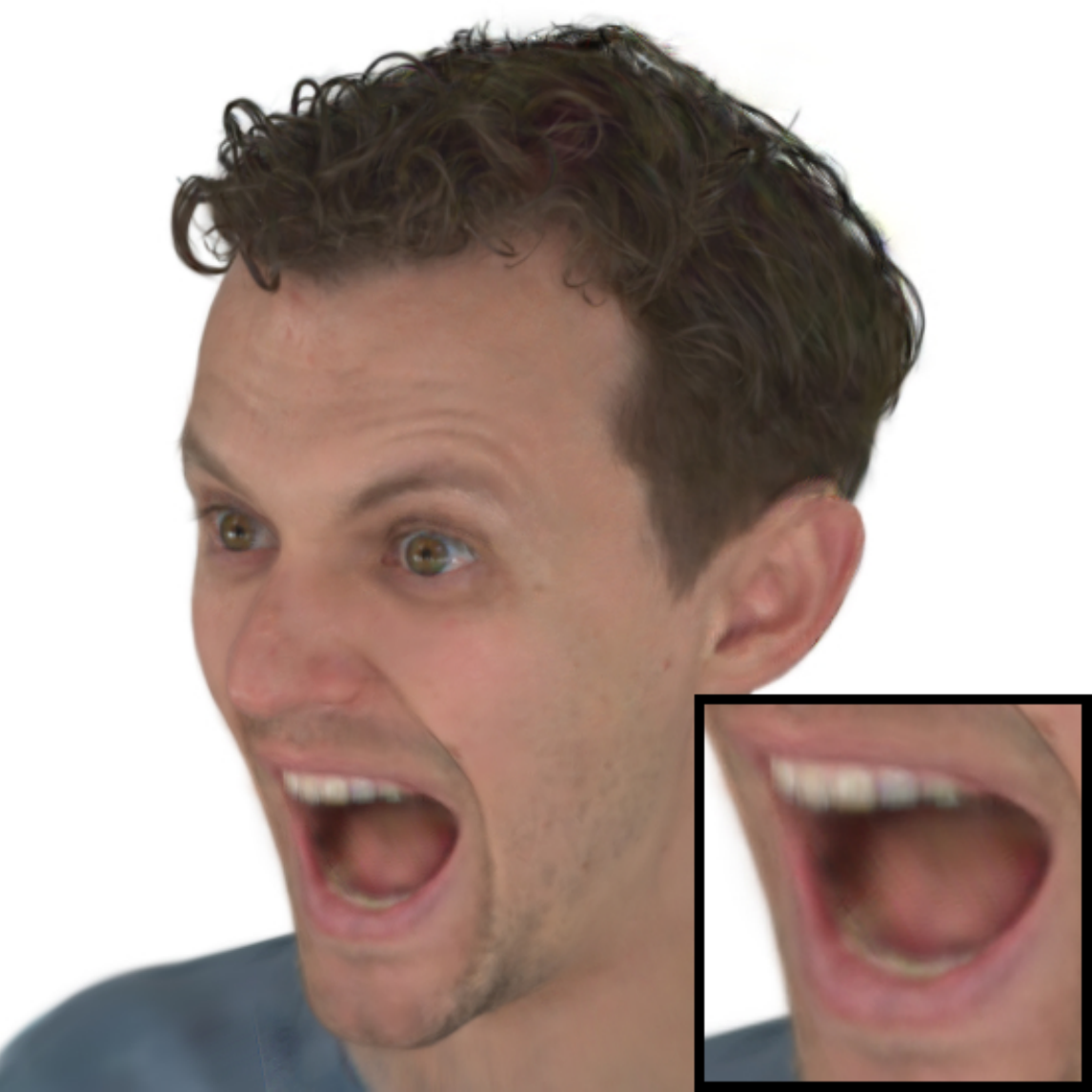}
        \caption{\name\ (Ours)}
    \end{subfigure}
    \begin{subfigure}{0.32\linewidth} 
        \includegraphics[width=\linewidth]{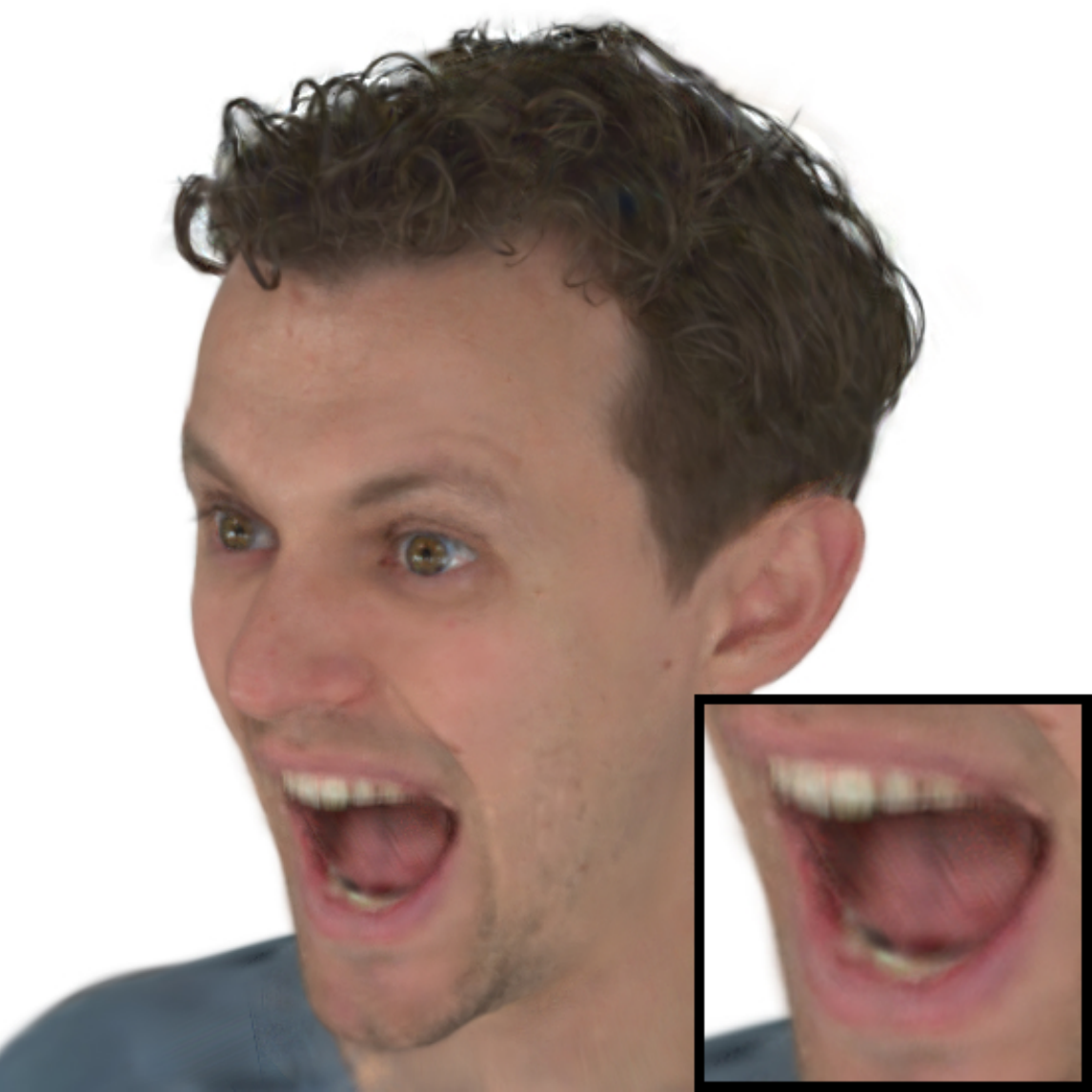}
        \caption{SVG-freeze}
    \end{subfigure}
    \begin{subfigure}{0.32\linewidth} 
        \includegraphics[width=\linewidth]{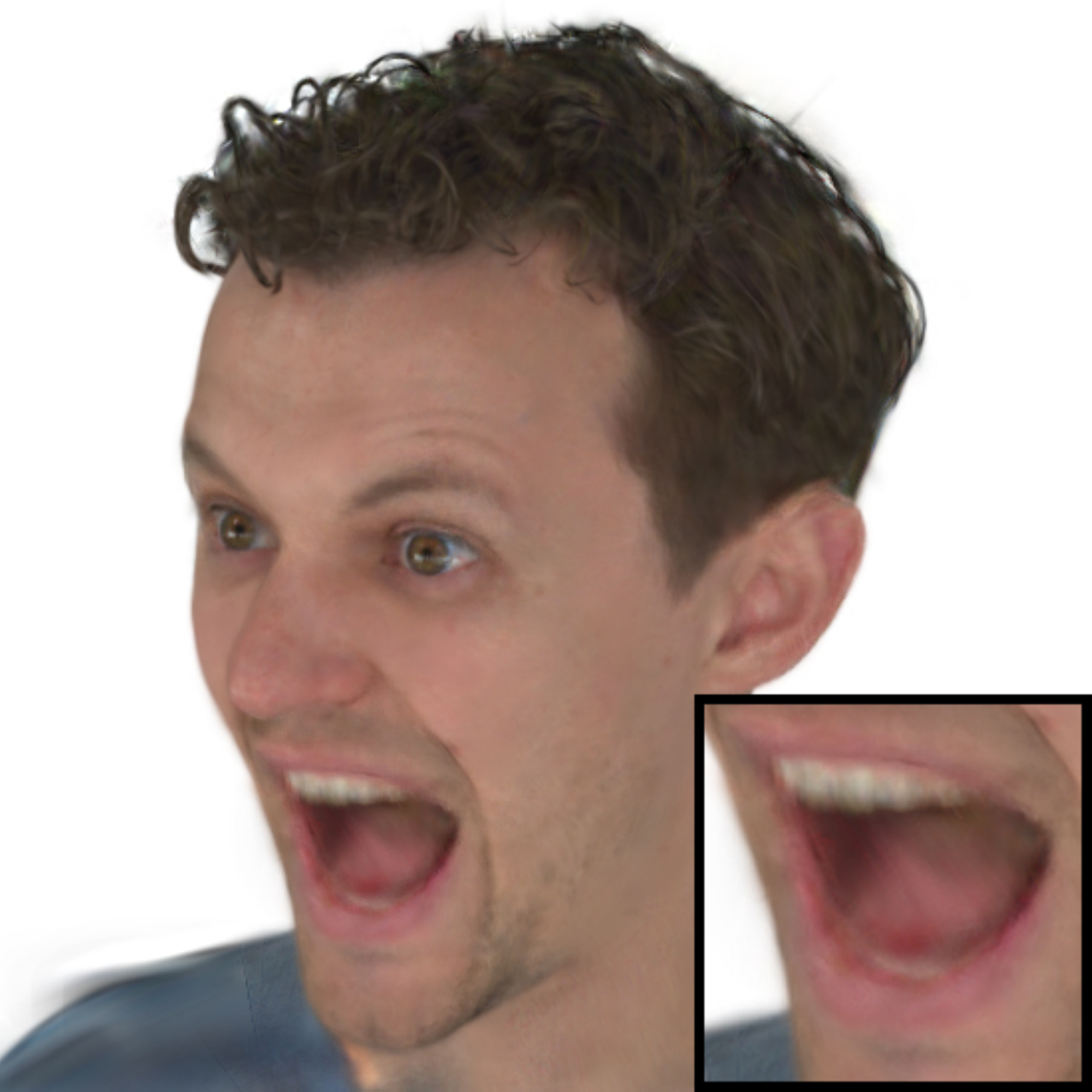}
        \caption{SVG-nofreeze}
    \end{subfigure}
    \begin{subfigure}{0.32\linewidth} 
        \includegraphics[width=\linewidth]{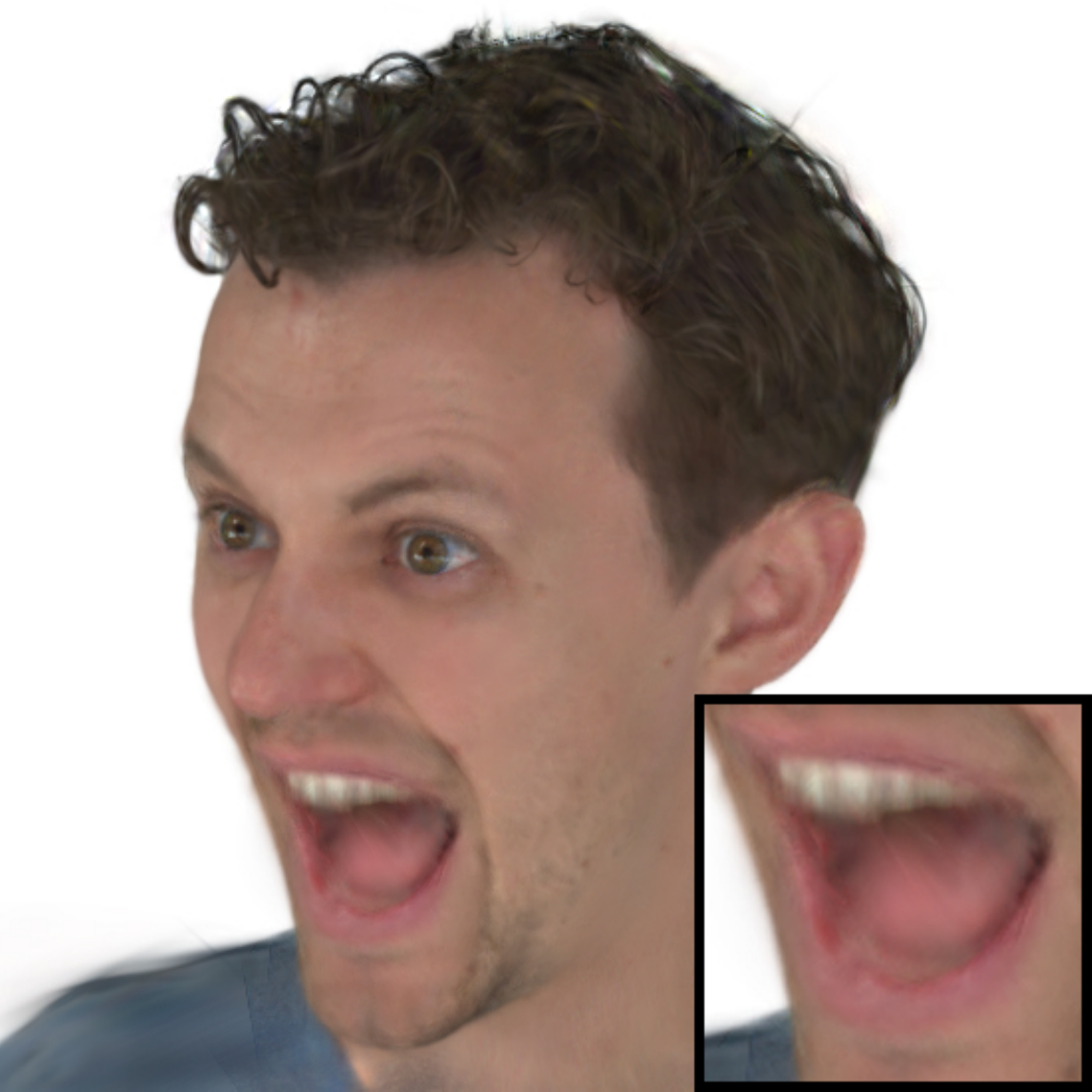}
        \caption{SVG-onestage}
    \end{subfigure}
    \vspace{-3mm}
    \caption{
    \textbf{Ablation Studies} on hierarchical optimization strategy. 
    Using alternative optimization strategies (c, d, e) decreases the renderings.
    }
    \vspace{-5mm}
    \label{fig:ablation-reconstruction}
\end{figure}

\vspace{\bfskip}
\noindent
\textbf{Hierarchical optimization strategy.}
We compare with some naive optimization strategy to verify the effectiveness of our optimization designs.
The quantitative results are reported in \cref{tab:ablation}(b.1)-(b.3) and the visual comparison are shown in ~\cref{fig:ablation-reconstruction}.
Jointly optimizing both types of Gaussians from scratch (\ie, SVG-onestage) generates unsatisfactory results (30.9 vs. 28.0 PSNR) due to the optimization is highly under-constrained.
``SVG-freeze'' and ``SVG-nofreeze'' attempt to freeze and unfreeze all parameters of the surf-GS during the joint optimization stage, respectively.
However, both of them struggle to obtain high-quality renderings (30.1/28.4 vs. 30.9 PSNR) compared to our selectively optimizing a subset of parameters (\ie, $o^{(s)}$ and $\mathcal{T}_\text{dy}$).
A possible reason is that optimizing only $o^{(s)}$ and $\mathcal{T}_\text{dy}$ is sufficient to support complementary modeling, and will not introduce too many optimization variables and cause under-constrained problems.

\vspace{\bfskip}
\noindent
\textbf{Loss functions.}
We investigate the effects of our proposed loss functions, and the quantitative results are in \cref{tab:ablation}(c.1)-(c.2).
While removing $\mathcal{L}_{\text{rgb}}^{\text{diff}}$ during the surface Gaussian optimization stage (\ie, SVG-nodirgb) does not reduce the evaluation metrics (31.1 vs. 30.9 PSNR), it results in a unreasonable texture image and impairs the editing quality (\cref{fig:ablation-edit}(e)).
``SVG-noalpha'' removes $\mathcal{L}_a$ during the joint optimization stage, causing worse rendering quality (30.2 vs. 30.9 PSNR) and transparent editing effects (\cref{fig:ablation-edit}(d)).

\subsection{Discussion on Mesh and Surface Gaussians}

To demonstrate the effectiveness of our surf-GS, we compare the reconstruction quality of surf-GS and mesh with VHAP~\cite{vhap-qian2024versatile}, as well as the quality achieved by combining surf-GS and mesh as first-stage results followed by Gaussians optimization, in comparison with HAHA~\cite{DBLP:conf/accv/SvitovMAB24}.

\begin{figure}[t]
    \centering
    \begin{subfigure}{0.24\linewidth}
        \includegraphics[width=\linewidth]{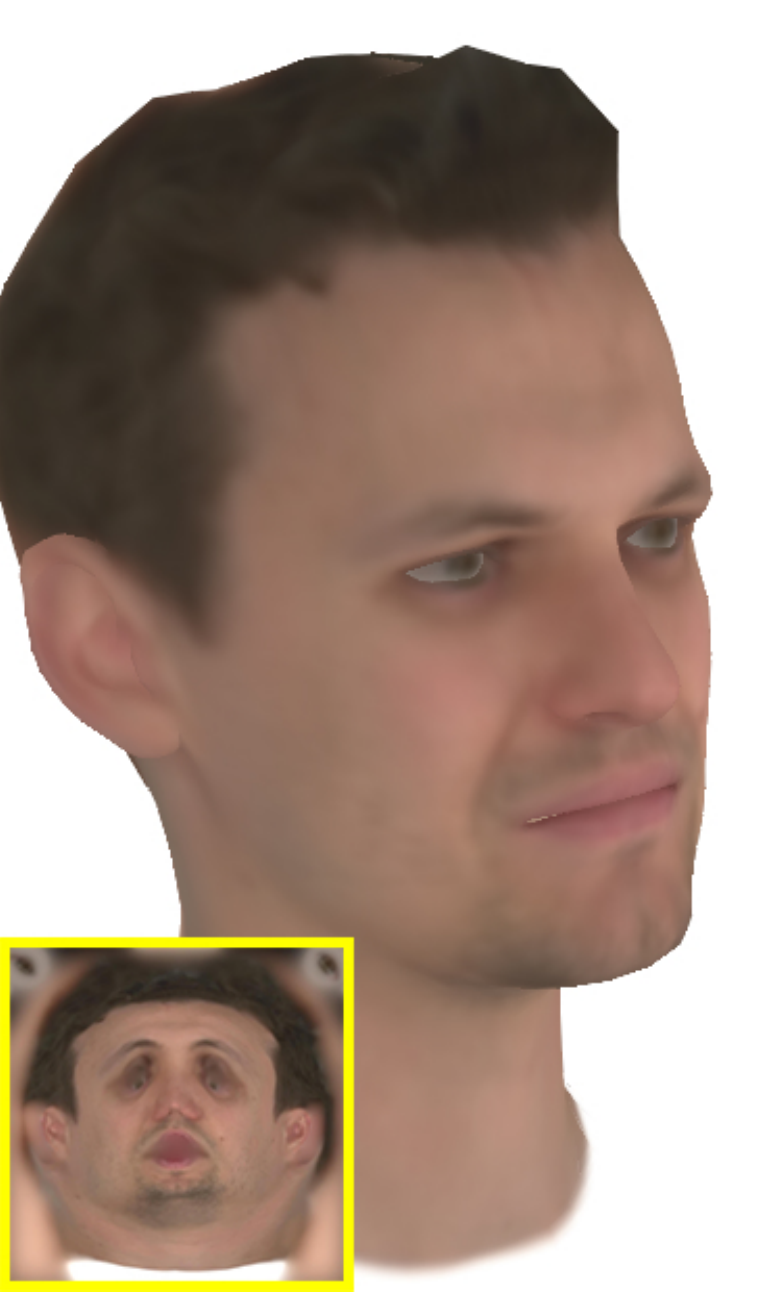}
        \caption{$\text{VHAP}_{\text{rgb}}$}
    \end{subfigure}
    \begin{subfigure}{0.24\linewidth}
        \includegraphics[width=\linewidth]{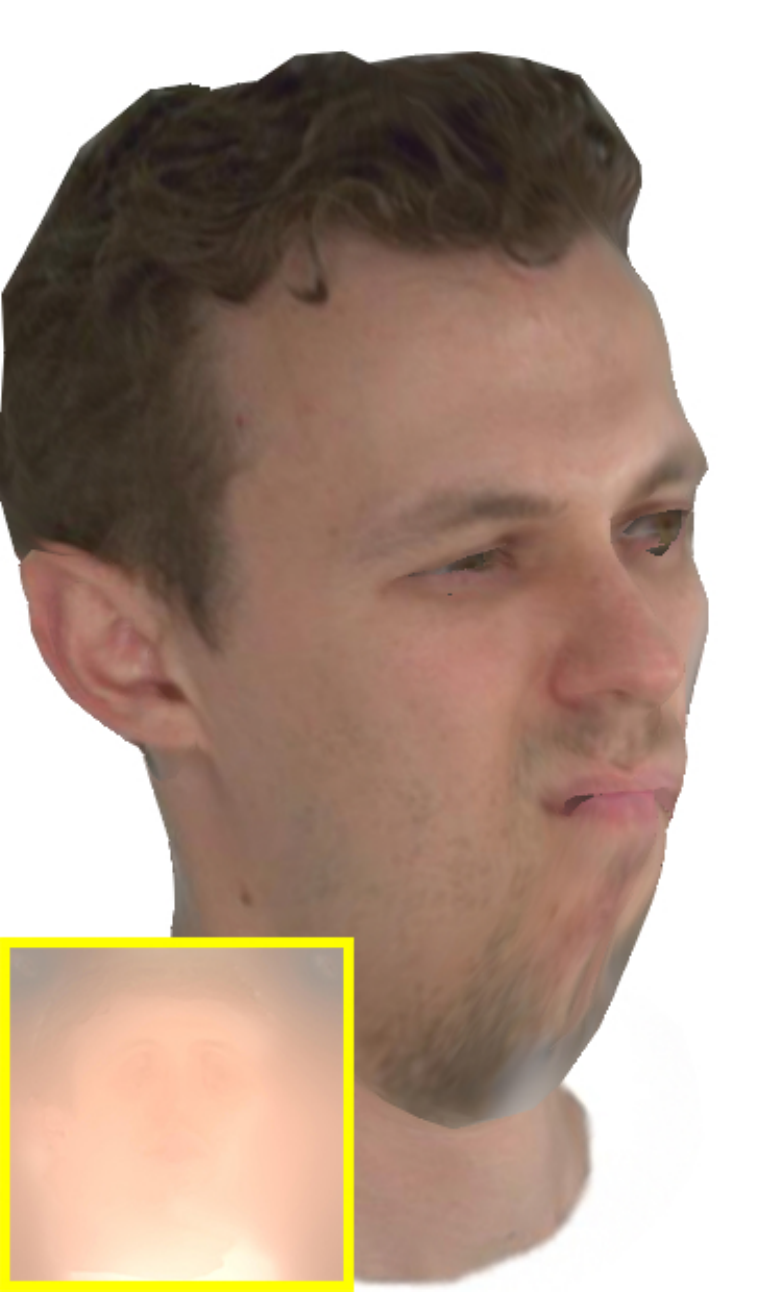}
        \caption{$\text{VHAP}_{\text{SH}}$}
    \end{subfigure}
    \begin{subfigure}{0.24\linewidth}
        \includegraphics[width=\linewidth]{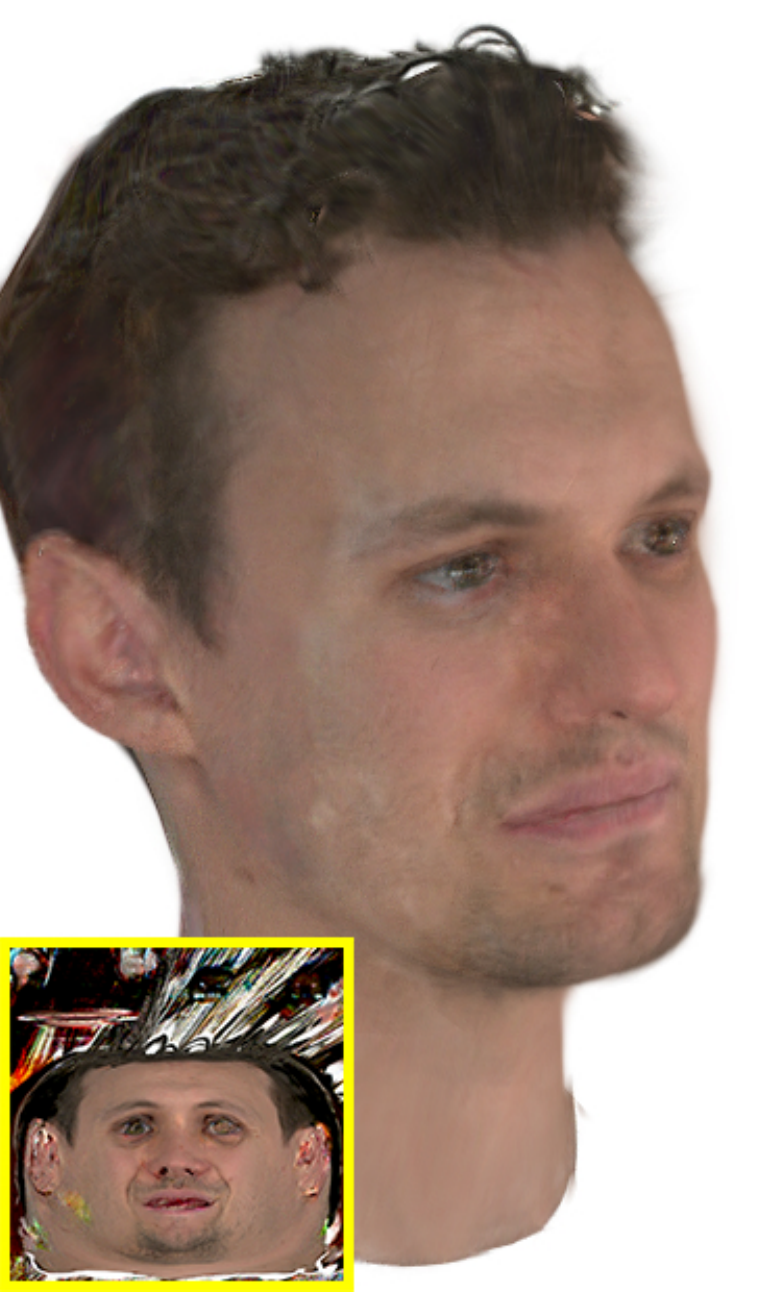}
        \caption{$\text{Surf-GS}_{\text{rgb}}^{\text{diff}}$}
    \end{subfigure}
    \begin{subfigure}{0.24\linewidth}
        \includegraphics[width=\linewidth]{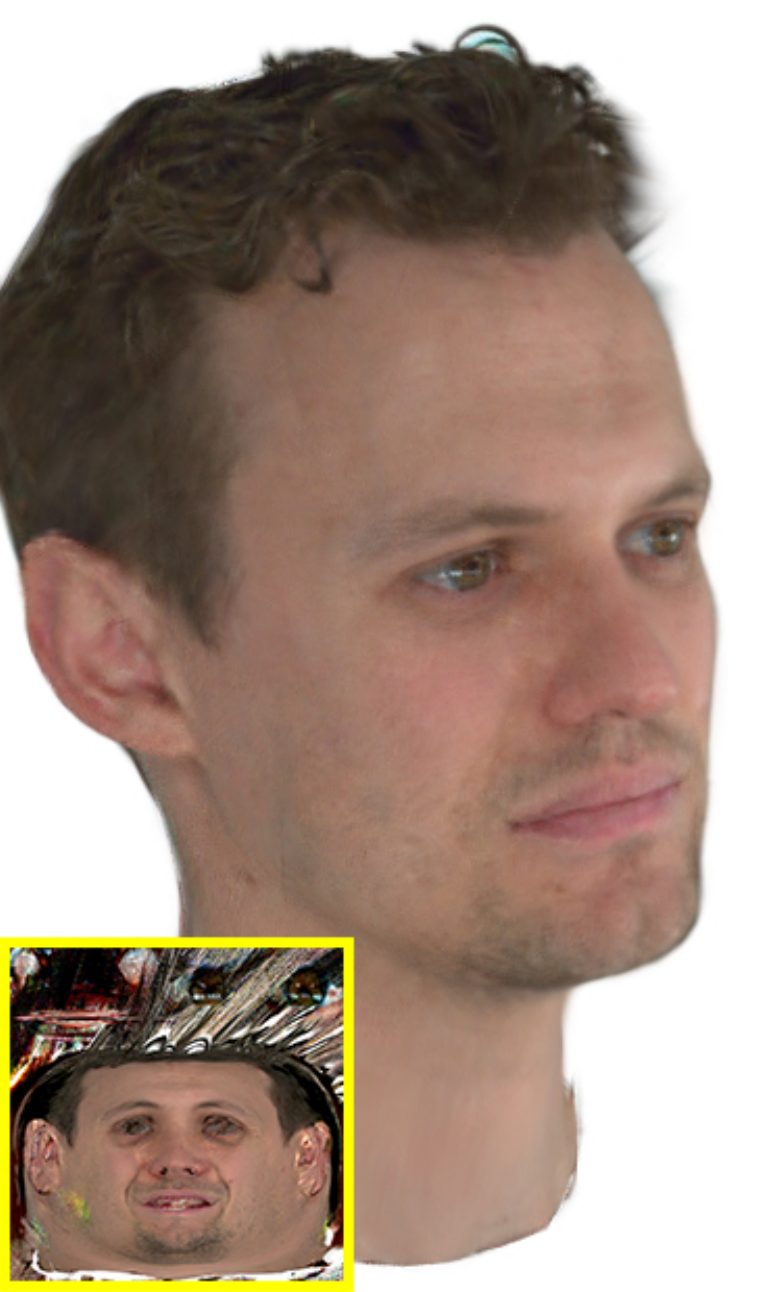}
        \caption{$\text{Surf-GS}_{\text{SH}}^{\text{diff}}$}
    \end{subfigure}
    \vspace{-2mm}
    \caption{
    \textbf{Comparisons between surface Gaussians and meshes.}
    Our surf-GS generates better renderings with higher-quality texture images.
    }
    \vspace{-4mm}
    \label{fig:vhap}
\end{figure}

As shown in \cref{tab:vhap}(d.1)-(d.2), due to the fixed topology, mesh-based optimization scheme with only RGB colors, \ie, $\text{VHAP}_{\text{rgb}}$, (no SH coefficients for view-dependent effects) struggles to obtain high-fidelity renderings compared to the same settings of surf-GS, \ie, $\text{Surf-GS}_{\text{rgb}}^{\text{diff}}$ (29.8 vs. 31.3 PSNR).
Moreover, surf-GS can easily introduce SH coefficients with a low cost (\ie, $\text{Surf-GS}_{\text{SH}}^{\text{diff}}$) for improved view-dependent effects (\cref{tab:vhap}(d.4)), while optimizing colored meshes with SH coefficients (\ie, $\text{VHAP}_{\text{SH}}^{\text{diff}}$) is a highly under-constrained problem which suffers from non-convergence (\cref{tab:vhap}(d.3)).
The visual comparisons are shown in \cref{fig:vhap}.

Due to space constraints, we include the detailed analysis and qualitative comparisons with HAHA in the supplementary material.

\begin{table}
  \centering
  \small
  \caption{
  \textbf{Comparisons between surface Gaussians and meshes.}
  Our surf-GS generates better renderings with higher-quality texture images, and can be extended with SH coefficients with a low cost for enhanced performance.
  }
  \vspace{-2mm}
      \resizebox{\linewidth}{!}{ 
        \begin{tabular}{l|l|ccccc} 
            \toprule
            Label & Name & PSNR$\uparrow$ & SSIM$\uparrow$ & LPIPS$\downarrow$ & Infer. Time & Memory \\
            \midrule
            (d.1) & $\text{VHAP}_{\text{rgb}}$ & 29.8 & 0.953 & 0.104 & $\sim$120fps & $\sim$1500MB \\
            (d.2) & $\text{Surf-GS}_{\text{rgb}}^{\text{diff}}$ & 31.3 & 0.953 & 0.043 & $\sim$75fps & $\sim$2000MB \\
            (d.3) & $\text{VHAP}_{\text{SH}}$ & 17.5 & 0.914 & 0.154 & $\sim$50fps & $\sim$14200MB \\
            (d.4) & $\text{Surf-GS}_{\text{SH}}^{\text{diff}}$ & 33.3 & 0.964 & 0.033 & $\sim$75fps & $\sim$2100MB \\
            \bottomrule
        \end{tabular}
      }
      \vspace{-6mm}
  \label{tab:vhap}
\end{table}
\vspace{-1mm}
\section{Conclusion}
\label{sec:conclusion}
In this paper, we present a novel Surface-Volumetric Gaussian representation for reconstructing high-fidelity, animatable and real-time editable head avatars, which adopts two types of Gaussians for complementary modeling. To enable fine-grained real-time appearance editing, surface Gaussians are introduced to disentangle the appearance of head avatars from geometry, which are bound on the mesh surface and leverage mesh-aware Gaussian UV mapping to fetch colors from learnable texture images (including diffuse and dynamic textures).
%
%
%
For enhanced rendering quality, we also design volumetric Gaussians that move around the mesh for complementarily modeling residual regions with fine-grained geometry(\eg, hair).
A differential hybrid rendering and hierarchical optimization strategy are further utilized to obtain the optimal performance.
%
In addition to generating high-fidelity renderings, our approach is the first to obtain explicit texture images for Gaussian head avatars, potentially supporting more applications.

\vspace{1.2mm}
\noindent
\textbf{Limitations.}
Although our approach is the first to obtain an explicit texture image for Gaussian head avatars and achieve high-quality renderings, there are still some limitations that deserve further exploration.
(1) Our surface Gaussians with explicit texture images struggles to accurately model non-Lambertian regions (e.g., hair), causing unsatisfactory effects when editing these regions.
(2) SVG-Head still can't achieve better rendering quality than methods that focus only on reconstruction.
Novel Gaussian UV mapping functions may be helpful for solving these issues.

\vspace{1.2mm}
\noindent

\clearpage

{
    \small
    \bibliographystyle{ieeenat_fullname}
    \bibliography{main}
}

\end{document}